\documentclass[pdflatex,sn-mathphys-num]{sn-jnl}

\usepackage{graphicx}
\usepackage{multirow}
\usepackage{amsmath,amssymb,amsfonts}
\usepackage{amsthm}
\usepackage{floatpag}
\usepackage{mathrsfs}
\usepackage[title]{appendix}
\usepackage{xcolor}
\usepackage{textcomp}
\usepackage{manyfoot}
\usepackage{booktabs}
\usepackage{algorithm}
\usepackage{algorithmicx}
\usepackage{algpseudocode}
\usepackage{listings}
\usepackage{array,tabularx,booktabs}
\usepackage{adjustbox}
\usepackage{subcaption}
\usepackage{soul, color}

\theoremstyle{thmstyleone}

\theoremstyle{thmstyletwo}

\theoremstyle{thmstylethree}

\begin{document}

\title[Neural Minimum-Distance Estimation for Collision-Aware Operation of Multi-Arm Laparoscopy Surgical Robots Through Learning-from-Simulation]{Neural Minimum-Distance Estimation for Collision-Aware Operation of Multi-Arm Laparoscopy Surgical Robots Through Learning-from-Simulation}

\author[1]{\fnm{Sarvin} \sur{Ghiasi}}
\author[2]{\fnm{Majid} \sur{Roshanfar}}
\author[1]{\fnm{Jake} \sur{Barralet}}
\author[1]{\fnm{Liane S.} \sur{Feldman}}
\author[1]{\fnm{Amir} \sur{Hooshiar}}\email{amir.hooshiar@mcgill.ca}

\affil[1]{\orgdiv{Surgical Performance Enhancement and Robotics (SuPER) Centre, Department of Surgery}, \orgname{McGill University}, \orgaddress{\city{Montreal}, \state{QC}, \country{Canada}}}
\affil[2]{\orgdiv{The Wilfred and Joyce Posluns Centre for Image Guided Innovation \& Therapeutic Intervention (PCIGITI)}, \orgname{The Hospital for Sick Children (SickKids)}, \orgaddress{\city{Toronto}, \state{ON}, \country{Canada}}}

\abstract{This study presents an integrated framework for enhancing the safety and operational efficiency of robotic arms in laparoscopic surgery by addressing minimum distance estimation between multi-arm manipulators and the associated collision-aware warning. By combining analytical modeling, real time simulation, and machine learning, the framework offers a robust solution for ensuring safe robotic operations. An analytical model was developed to estimate the minimum distances between robotic arms based on their joint configurations, offering theoretical calculations that serve as both a validation tool and a benchmark. To complement this, a 3D simulation environment was created to model two 7 DOF Kinova robotic arms (Kinova inc., Boisbriand, QC, Canada), generating a diverse dataset of configurations for distance estimation and collision warning. Using these insights, a deep residual neural network model was trained with joint configurations as inputs. On the held out validation set, the model achieves $R^2 = 0.940$, RMSE $= 42.0$~mm, MAE $= 28.7$~mm, and a near zero mean bias, demonstrating strong predictive accuracy and consistent generalization across the workspace. The framework is intended as an early collision warning layer, where a warning is triggered when the predicted inter-arm distance falls below a 0.2~m threshold, which corresponds to a surface to surface clearance of approximately 50~mm given the Kinova Gen3 (Kinova inc., Boisbriand, QC, Canada) cross sectional radius. This work demonstrates the effectiveness of combining analytical modeling with machine learning to enhance the precision and reliability of multi-arm robotic systems.}


\maketitle

\section{Introduction}
\label{Introduction}
\subsection{Background}
\label{Background}
Minimally invasive techniques such as laparoscopy have transformed modern surgery, improving safety and patient outcomes~\cite{jeganathan2025minimally}. However, laparoscopic procedures remain limited by restricted instrument dexterity, two-dimensional visualization, and suboptimal surgeon ergonomics. Robotic systems have emerged to address these challenges by enhancing precision, range of motion, and control, thereby overcoming key limitations of conventional laparoscopy~\cite{jung2015robotic,cochetti2020combined,kwon2022comparison, roshanfar2025advanced}. The advanced design of robotic surgical systems plays a crucial role in reducing the physical strain and operational difficulties associated with traditional laparoscopy. As demonstrated by Thomas et~al., the Versius Surgical System (CMR Surgical Ltd, Cambridge, United Kingdom) replicates the articulation of the human arm, offering improved surgical access and maneuverability compared to standard laparoscopic tools~\cite{thomas2021preclinical}. Similarly, robotic surgery has been shown to provide a more effective alternative by mitigating ergonomic and operational challenges faced by surgeons~\cite{wehrmann2023clinical}. The work of Soumpasis et~al.~\cite{soumpasis2023safe} further emphasizes that robotic systems enhance flexibility, visualization, and access, directly addressing the limitations of conventional instruments. According to Silva et~al.~\cite{silva2018introduction}, these systems can significantly reduce surgeon fatigue while providing intuitive instrument control. This innovation not only improves procedural efficiency but also accelerates the training process for novice surgeons, enabling them to acquire complex surgical skills more quickly. Ruzzenente et~al.~\cite{ruzzenente2020robotic} reported that the learning curve for robotic surgery is generally shorter than that for laparoscopy, particularly for surgeons already proficient in open procedures. Likewise, robotic systems facilitate the transfer of skills between robotic-assisted and laparoscopic surgery, simplifying training and skill retention~\cite{hardon2023crossover,schmidt2024laparoscopic}.  

\begin{figure*}
\centering
\begin{tabular}{cc}
\includegraphics[width=0.49\textwidth]{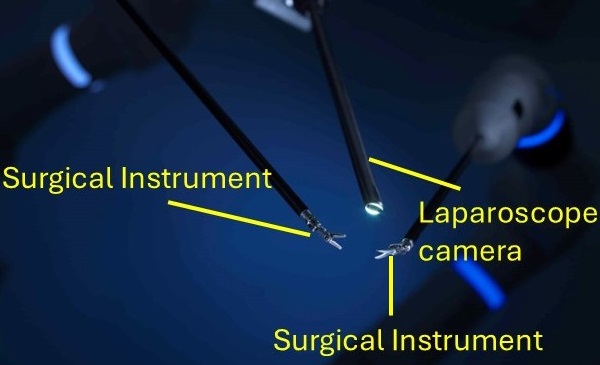} &
\includegraphics[width=0.46\textwidth]{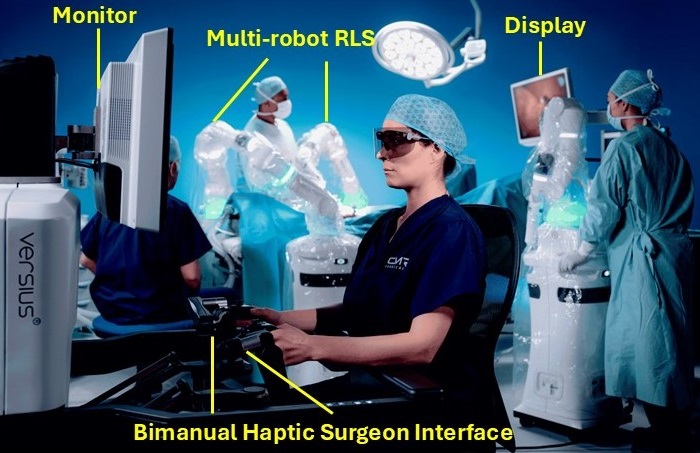} \\
(a) & (b)
\end{tabular}
\caption{(a) Robot-controlled laparoscopic instruments with an integrated laparoscope camera providing a magnified operative view. (b) A surgeon operating through a bimanual haptic interface within a multi-robot robotic laparoscopic system (RLS), with real-time visualization displayed on the monitor and overhead surgical display. These images illustrate the typical setup of a robot-assisted laparoscopic surgery environment, demonstrating both the operative field and the surgeon console. Images courtesy of CMR Surgical (Versius Surgical System).}
\label{robotic_surgery}
\end{figure*}

Beyond ergonomic and training advantages, robotic surgery has also been associated with improved postoperative outcomes. Studies have shown that patients undergoing robotic-assisted procedures experience fewer complications and shorter hospital stays than those treated with conventional laparoscopy~\cite{roh2020comparison}. Gerdes et~al.~\cite{gerdes2022results} reported superior postoperative recovery, including reduced pain and faster rehabilitation, while Chen et~al.~\cite{chen2022vinci} associated robotic approaches with enhanced recovery protocols. These findings highlight the potential of robotic systems to optimize patient outcomes, particularly in gynecological procedures~\cite{manchanda2024comprehensive}. Moreover, robotic surgery holds significant value in surgical education, where the complexity of laparoscopic techniques poses challenges for trainees. By providing a stable camera platform and advanced 3D visualization, robotic systems improve spatial awareness and facilitate skill acquisition, as illustrated in Fig.~\ref{robotic_surgery}.

\subsection{Clinical Motivation for Collision-Aware Robotic Systems}
\label{Clinical Motivation for Collision-Aware Robotic Systems}
Robotic-assisted surgery enhances precision and ergonomics by providing improved dexterity, three-dimensional visualization, and tremor filtration, which together lead to better outcomes and reduced fatigue for surgeons~\cite{li2018robotic,li2019distal}. Despite these advantages, challenges persist that can compromise safety and efficiency, particularly the risk of instrument collisions and workspace constraints~\cite{montagne2021robotic}. Sun et~al.~\cite{sun2020visual} reported that coordinating multiple robotic arms within confined surgical environments often leads to reduced maneuverability and unintentional collisions, potentially causing tissue damage or mechanical interference. Similarly, Abiri et~al.~\cite{abiri2017visual} emphasized the need for optimizing robotic system design to improve flexibility and reduce the likelihood of tissue trauma, while Kim et~al.~\cite{kim2016development} identified arm misalignment as a frequent source of accidental contact during robotic procedures.  

Unintended collisions remain a critical yet underexplored challenge across several minimally invasive domains, including laparoscopic, endonasal, and microsurgical procedures~\cite{ueda2017toward}. In these operations, surgeons often face limited visibility near the instrument tip, increasing the likelihood of inadvertent contact with surrounding tissues or instruments. Kelly et~al.~\cite{kelly2020single} and Nigicser et~al.~\cite{nigicser2022magnetic} both highlighted that restricted visibility and workspace constraints hinder precise instrument coordination in multi-arm robotic surgery. Mendelsohn et~al.~\cite{mendelsohn2020transoral} similarly noted that limited visual access can result in unexpected tool collisions in transoral procedures. In microsurgery, spatial restrictions further limit dexterity and elevate procedural complexity~\cite{marinho2019dynamic}, while Yoshimura et~al.~\cite{yoshimura2020single} observed that collision zones near the tool tip often fall outside the surgeon’s visual field. Such limitations are especially consequential in high-precision interventions, where even minor contact can jeopardize surgical accuracy and patient safety~\cite{minig2017minimally}. These challenges underscore the pressing need for reliable, real-time collision detection and avoidance systems to enhance safety and performance in robotic-assisted surgery~\cite{he2019human, roshanfar2022stiffness}.

\subsection{Prior Work}
\label{related_works}
Accurate detection and prevention of collisions in robotic systems require an understanding of workspace geometry and often rely on computationally intensive motion-planning algorithms~\cite{jimenez2005collision}. To address this challenge, researchers have proposed a range of sensing and modeling approaches to enhance safety and control in robotic-assisted environments. Cirillo et~al.~\cite{cirillo2015conformable} developed a sensorized flexible skin capable of detecting contact position and applied forces for managing both intentional and unintentional human-robot interactions. While promising, the method’s dependence on a single sensing modality limits adaptability in dynamic environments and warrants further comparison with alternative force-sensing approaches, such as torque or residual-based methods. Marinho et~al.~\cite{marinho2019dynamic} introduced a vector-field-inequalities approach for autonomous motion guidance, minimizing the risk of collisions during surgical tasks. However, the method’s reliance on Jacobian-based modeling restricts its ability to capture nonlinear dynamics in complex procedures. Other studies employed virtual fixtures as safety boundaries to prevent instruments from entering restricted zones~\cite{vitrani2016applying,marinho2020smartarm}. Although effective, these approaches face challenges related to the fulcrum constraint, which introduces multiple possible solutions for replicating distal forces during proximal co-manipulation.  

Subsequent work extended these concepts to improve teleoperation and safety. Marinho et~al.~\cite{marinho2019unified} proposed a unified vector-field framework for smooth teleoperation and collision avoidance, though its performance in flexible or dynamic geometries remains unverified. Similarly, Marinho et~al.~\cite{marinho2020virtual} applied guidance virtual fixtures—such as looping trajectories and cylindrical constraints—to enhance precision and safety, but noted minimal improvement in task completion time. Xin et~al.~\cite{xin6research} integrated virtual fixtures with impedance-conductance control to prevent human-robot co-surgery errors; however, experiments were limited to two-dimensional settings, reducing applicability to real-world three-dimensional surgical environments. Lin et~al.~\cite{lin2019virtual} further employed a human-arm kinematic model to construct guided virtual fixtures that restrict robot motion during dragging, though validation was confined to simulation studies. Beyond virtual fixtures, path-planning and control strategies have been explored for safe human-robot collaboration. Wang et~al.~\cite{wang2023path} reduced oscillations in robot trajectories using velocity potential fields, achieving real-time performance but focusing only on single-arm manipulators. Herbster et~al.~\cite{herbster2023modeling} modeled contact forces in constrained human-robot collisions to advance collaborative robot (cobot) safety, though the model’s generalizability is limited by simplified contact mechanics and testing across a few platforms.  

Zhu et~al.~\cite{zhu2012model} proposed a geometric collision detection method for a maxillofacial multi-arm surgical robot (MMSR) using cylindrical and spherical primitives to represent arms and obstacles. While computationally efficient, the simplified geometry limits applicability in anatomically complex surgical spaces. Hao et~al.~\cite{hao2021application} presented an adaptive genetic algorithm based on collision detection (AGACD) that improved convergence and computation time in planning tasks, though the approach neglected mechanical constraints and dynamic environments. In parallel, sensorless detection techniques have gained attention for reducing hardware complexity. Lee et~al.~\cite{lee2015sensorless} and Li et~al.~\cite{li2019virtual} developed sensorless collision detection systems that identified external torques using friction modeling and joint motor currents. While these methods avoid costly sensors, they remain less accurate than traditional sensor-based systems and require refinement of friction models for improved precision. These limitations highlight the need for lightweight, accurate, and real-time collision detection frameworks capable of adapting to the constrained and dynamic conditions of surgical workspaces. A summary of some of the related studies are presented in Table~\ref{tab:review}.

\begin{table*}
\centering
\caption{Summary of related studies on collision detection and avoidance methods.}
\label{tab:review}
\begin{adjustbox}{width=\textwidth}
\begin{tabular}{|
m{0.6cm}<{\raggedright\arraybackslash}|
m{3.2cm}<{\raggedright\arraybackslash}|
m{3.6cm}<{\raggedright\arraybackslash}|
m{3.6cm}<{\raggedright\arraybackslash}|
m{3.6cm}<{\raggedright\arraybackslash}|
}\hline
\textbf{Ref.} & \textbf{Application} & \textbf{Method} & \textbf{Advantages} & \textbf{Limitations} \\
\hline
\cite{cirillo2015conformable} & Human-robot interaction & Sensorized skin & Measures contact directly; adaptable & Single-sensor reliance; needs comparative studies \\
\hline
\cite{marinho2019dynamic} & Safe guidance in robotic surgery & Vector-field inequalities & Reduces collision risks & Jacobian model limitations \\
\hline
\cite{vitrani2016applying} & Enforcing safe zones & Virtual fixtures & Prevents unsafe movement & Fulcrum constraints \\
\hline
\cite{marinho2019unified} & Teleoperation & Unified vector-field framework & Smooth teleoperation; collision avoidance & Unverified for dynamic settings \\
\hline
\cite{xin6research} & Human-machine co-surgery & Impedance control with virtual fixtures & Prevents errors; integrates force feedback & 2D testing only \\
\hline
\cite{lin2019virtual} & Guided movement restriction & Human arm kinematics & Enhances precision; simulation-tested & No real-world validation \\
\hline
\cite{zhu2012model} & Self-collision detection for MMSR & Geometric modeling & Computational efficiency & Limited for complex shapes \\
\hline
\cite{hao2021application} & Path planning & Adaptive Genetic Algorithm & Resolves convergence issues & Ignores mechanical constraints \\
\hline
\cite{lee2015sensorless} & Collision detection in human-robot tasks & Sensorless (friction/current) & Cost-effective; no sensors needed & Requires model refinement \\
\hline
\end{tabular}
\end{adjustbox}
\end{table*}

\subsection{Contributions}
\label{contributions}
To address the challenges of collision detection and workspace safety in multi-arm robotic systems, this study presents a comprehensive framework that integrates analytical modeling, real-time simulation, and learning-based prediction. A novel analytical algorithm is introduced for modeling robotic arms through a multi-linear shape representation, allowing efficient estimation of inter-arm distances and detection of potential collisions during operation. The proposed method was validated through real-time experiments across multiple robotic configurations and further examined within a custom 3D simulation environment developed to visualize and analyze dynamic system behavior. In addition, a deep neural network (NN) was trained and validated on a large dataset of simulated configurations covering a wide range of surgical scenarios. Comparative analyses between the analytical model, simulation outcomes, and experimental measurements demonstrated strong agreement, confirming the robustness and accuracy of the proposed framework for safe and reliable robotic operation. The main contributions of this work are summarized as follows:
\begin{enumerate}
    \item Development and validation of a robust analytical geometric framework for accurately predicting the minimum distance between two serial robotic arms.  
    \item Design and implementation of a real-time simulation platform for visualizing and validating minimum-distance estimation between robotic arms under varying configurations.  
    \item Training and deployment of a NN model for fast and accurate real-time prediction of minimum inter-arm distances in multi-arm robotic systems.  
\end{enumerate}

\section{Materials and Methods}
\label{Materials and Methods}
\subsection{Collision Detection Framework}
\label{Collision Detection Framework}
This study proposes a collision detection framework that integrates simulation-based learning, analytical modeling, and deep NN prediction to enhance the safety and situational awareness of multi-arm robotic systems. The framework follows a learning from simulation approach, combining large-scale synthetic data generation with real-time inference for collision prediction and warning. Simulation-driven optimization with task-driven constraints has also been used in surgical robotics to improve safe and robust operation, motivating the use of virtual environments for systematic design and validation~\cite{roshanfar2024design}. In the first stage, a physics-based simulation environment was developed in Unity to model two 7-DOF robotic arms across a wide range of spatial configurations. Rigid serial-link manipulators were chosen because the clinically deployed multi-arm robotic laparoscopic platforms that motivate this work, including the Intuitive da Vinci and the CMR Versius systems~\cite{thomas2021preclinical,wehrmann2023clinical,soumpasis2023safe}, are built on this architecture, and inter-arm interference between the bulky proximal links of adjacent arms remains a primary safety concern in such systems. The Kinova Gen3 was selected as a research proxy for this class of platforms because it provides a representative 7-DOF serial kinematics with a reach of approximately 0.9~m and kinematic redundancy sufficient for the constrained-workspace dexterity encountered in laparoscopic teleoperation; its workspace is therefore not a limiting factor for the inter-arm collision problem considered here, which concerns the macro-scale manipulator bodies outside the patient rather than the distal tool motion inside the cavity. Continuum manipulators are a complementary class of devices that primarily target intraluminal and tortuous-path interventions, where the dominant safety problem is tissue contact rather than interference between external links, and they therefore address a different clinical scenario from the one studied here. The proposed framework nevertheless relies solely on forward kinematics and is platform-agnostic at the link level, so it can be extended to continuum manipulators by replacing the B\'ezier link representation with a shape-parameterized backbone model, which is identified as a direction for future work.

\begin{figure*}
    \centering
    \includegraphics[width=\textwidth]{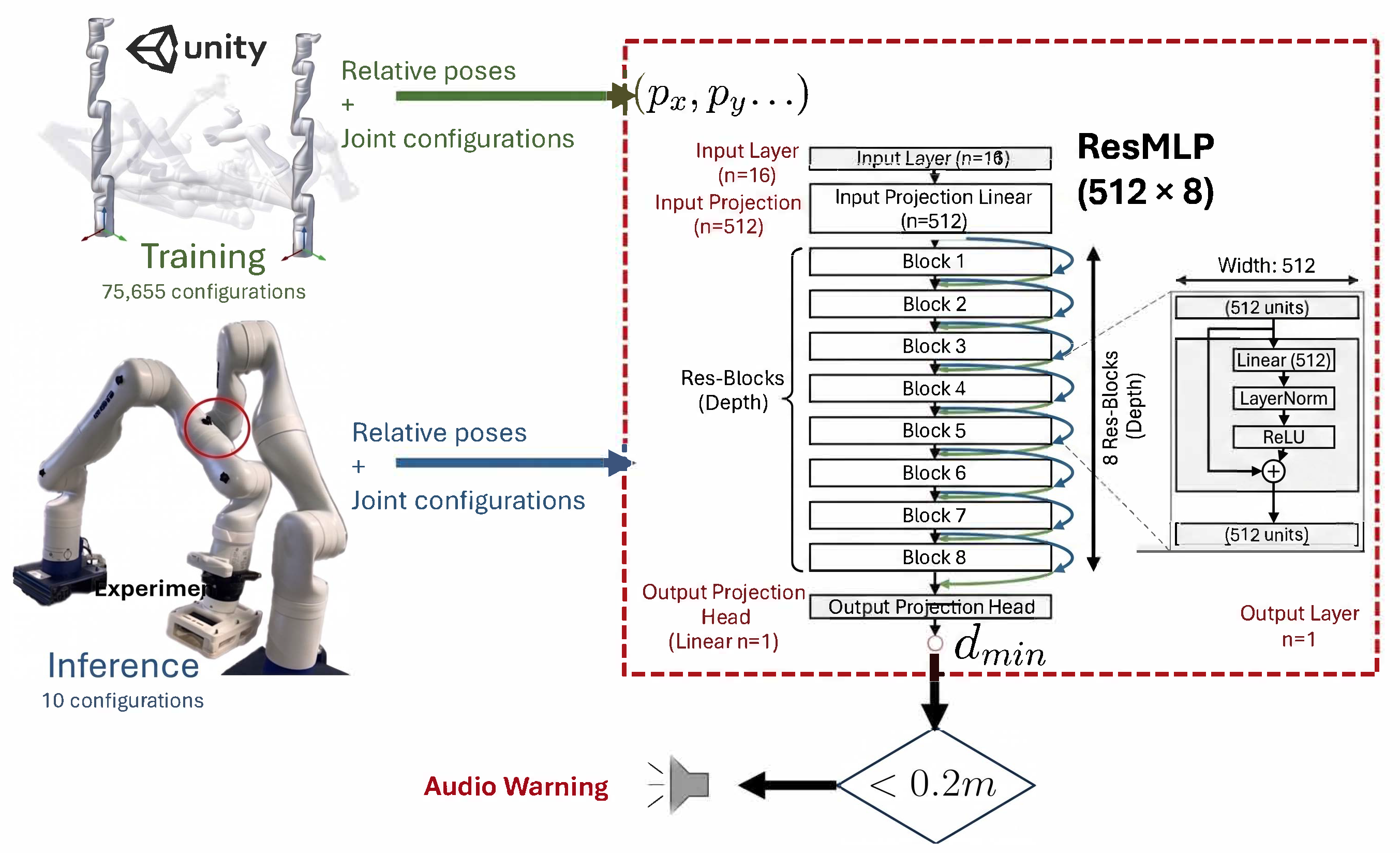}
    \caption{Overview of the proposed learning from simulation collision detection framework. The Unity-based simulation generates 75,655 robot configurations for training. A deep NN takes the eight selected joint angles as inputs and predicts the minimum distance (\(d_{min}\)) between the arms during inference on real robotic setups, issuing an audio warning when \(d_{min} < 0.2~\text{m}\). Blue blocks denote the simulation data-generation stage, green blocks the neural network training and inference stage, and the red marker the triggered collision warning.}
    \label{fig:framework}
\end{figure*}

The simulation platform enabled the generation of 75,655 random configurations by varying joint angles and relative poses between the robots. An analytical proximity estimation algorithm was embedded into the simulator to compute the minimum distance (\(d_{min}\)) between the arms for each configuration. The resulting dataset, consisting of joint configurations and relative pose parameters, was used to train a deep NN designed to predict \(d_{min}\) in real time. During the inference phase, experimental configurations obtained from two real Kinova Gen3 robotic arms were used as input to the trained network. The model receives the joint angles of the two robots and predicts the minimum inter-arm distance. When the predicted \(d_{min}\) falls below 0.2~m, the system triggers an audio warning to alert the operator of a potential collision. As detailed in Section~\ref{Performance Metrics}, this threshold is anchored in the Kinova Gen3 geometry: with an approximate cross sectional radius of 75~mm, a center line to center line distance of 150~mm corresponds to surface to surface contact, so the 0.2~m threshold provides an additional clearance margin of approximately 50~mm that allows the teleoperation system to arrest motion before the arms are driven into collision. The overall learning from simulation workflow, including dataset generation, NN architecture, and real-time inference, is illustrated in Fig.~\ref{fig:framework}.  

\subsection{Kinematic Modeling of the Robotic System}
\label{Multi-robot configuration}
\begin{figure*}
    \centering
    \includegraphics[width=\textwidth]{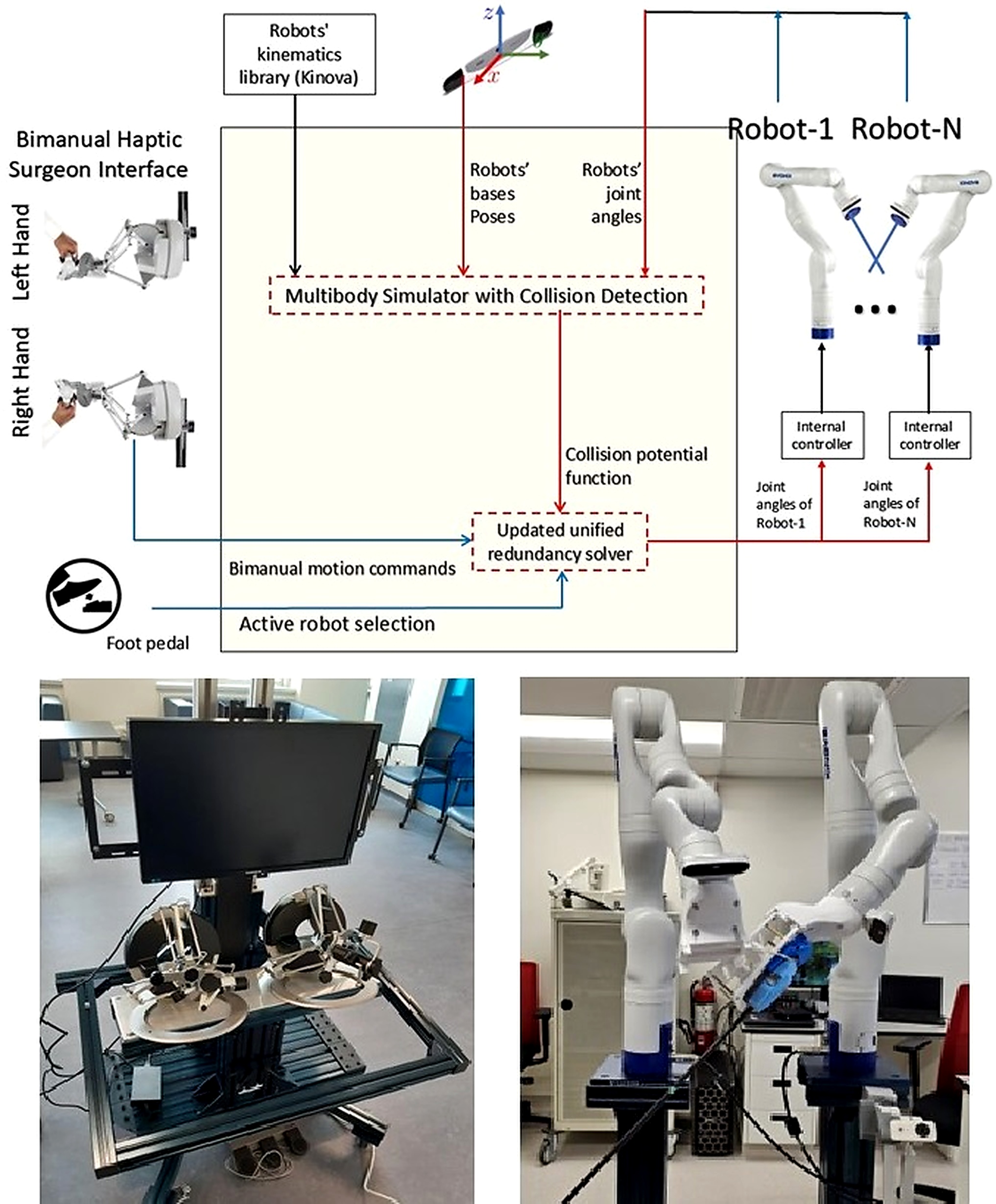}
    \caption{Overview of the experimental setup used in this research. The configuration includes: (left) a bimanual haptic surgeon interface for controlling both robotic arms with left and right hands and a foot pedal for robot selection; (center) a multibody simulator that integrates robot kinematics, collision detection, and redundancy resolution; and (right) dual Kinova Gen3 collaborative robotic arms performing coordinated manipulation tasks.}
    \label{fig:setup-schematic}
\end{figure*}

\begin{figure*}
    \centering
    \begin{subfigure}[t]{0.58\textwidth}
        \centering
        \includegraphics[width=\textwidth]{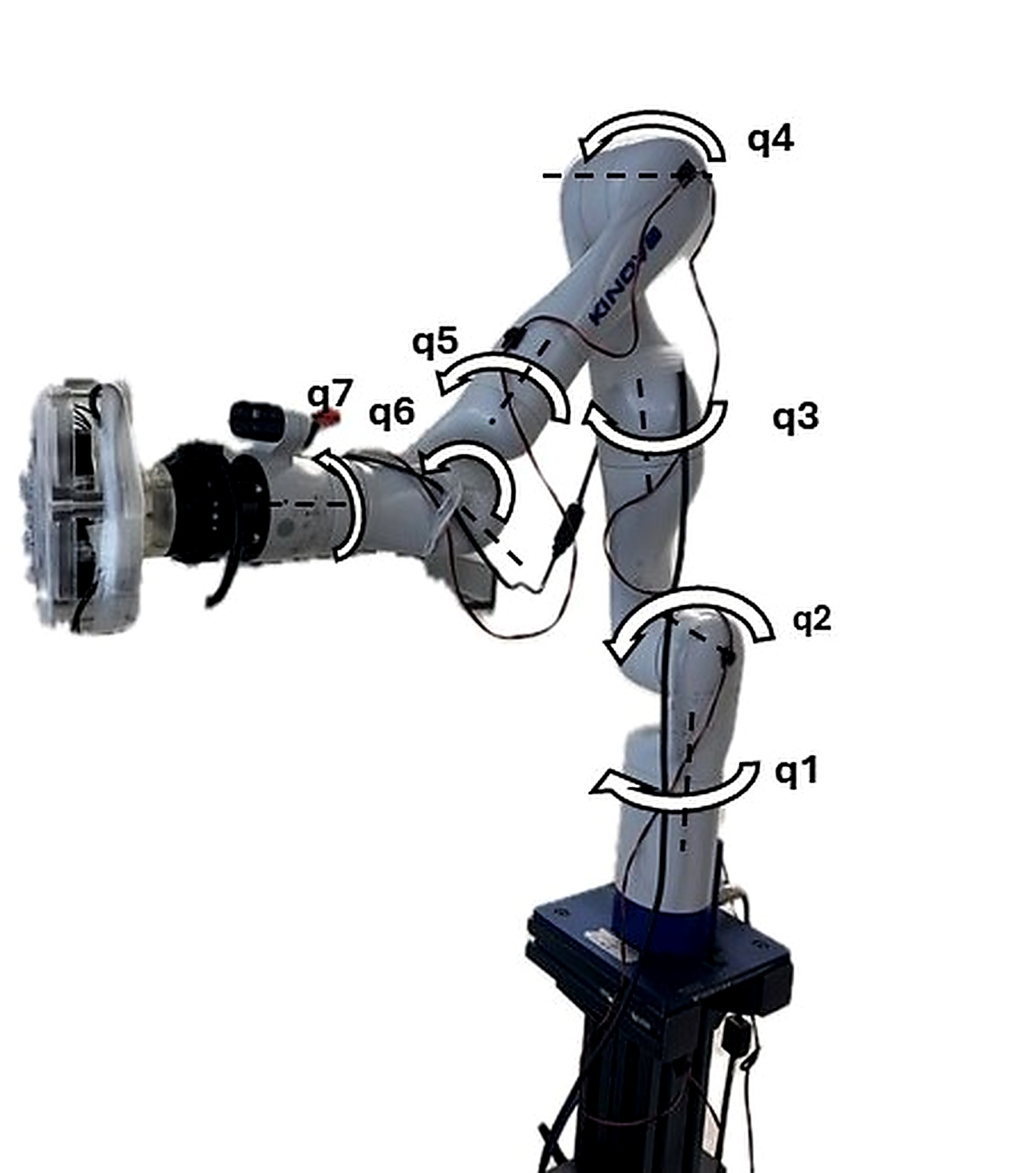}
        \caption{Joint rotations}
        \label{fig:single-arm}
    \end{subfigure}
    \hfill
    \begin{subfigure}[t]{0.40\textwidth}
        \centering
        \includegraphics[width=\textwidth]{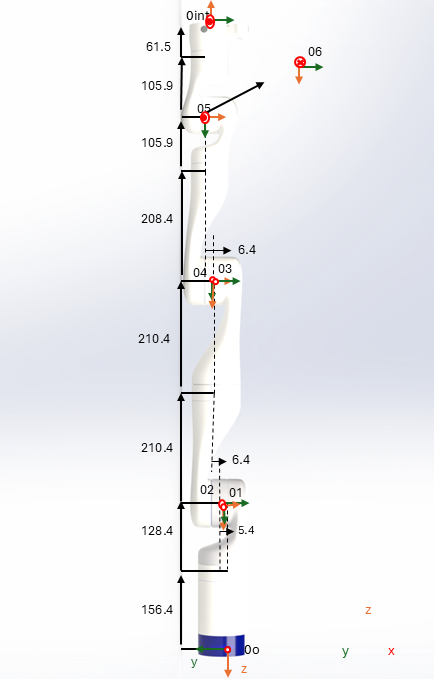}
        \caption{DH frames}
        \label{fig:kinova-rendered}
    \end{subfigure}
    \caption{Modeling of the Kinova Gen3 7-DOF robotic arm. 
    (a) Rotation of joint angles illustrating the seven revolute joints (\(q_1\) to \(q_7\)) that define the manipulator’s full configuration and contribute to the positioning and orientation of the end effector. 
    (b) Denavit-Hartenberg (DH) frame definitions showing coordinate frame assignments for each joint according to the DH convention, representing the spatial orientation and relative displacement between adjacent links.}
    \label{fig:kinova_combined}
\end{figure*}

The overall schematic of the experimental setup used in this study is shown in Fig.~\ref{fig:setup-schematic}. The system consists of a bimanual haptic surgeon interface, a multibody simulator with integrated collision detection, and multiple collaborative robotic manipulators. The bimanual interface enables intuitive control of the dual robotic arms using left and right hand devices, while a foot pedal allows the user to actively select which robot to operate. The simulator receives motion commands from the haptic interface and computes the corresponding kinematic parameters and joint angles for each robotic arm. It provides a comprehensive model of the robots’ motion by calculating and visualizing their positions, velocities, and accelerations based on the input joint configurations. The simulator also includes a collision detection module that continuously monitors potential interactions between the robotic arms and surrounding objects within the workspace. A redundancy resolution algorithm is implemented to determine optimal joint configurations that achieve the desired end-effector trajectory while avoiding singularities or workspace conflicts. Internal control mechanisms ensure precise execution of motion commands while enforcing joint limits, torque constraints, and synchronized movement between the robots. Furthermore, the simulator delivers haptic feedback cues through the interface, allowing the operator to feel interaction forces and improving awareness of the robots’ behavior during operation. The integration of kinematic modeling, collision detection, redundancy resolution, and haptic feedback supports the safe and accurate assessment of robotic motion in complex surgical environments. A forward kinematic model was developed using the Denavit-Hartenberg (DH) convention to describe the spatial configuration of each robot. As shown in Fig.~\ref{fig:single-arm}, each robotic arm includes seven revolute joints (\(i \in 1, 2, \ldots, 7\)) arranged in a spherical wrist configuration. The naming convention and corresponding axes of rotation are depicted in Fig.~\ref{fig:kinova-rendered}. The coordinate frames for each joint were defined according to the Kinova Gen3 Ultra Lightweight Robot User Guide. The DH parameters including link length (\(a_i\)), twist angle (\(\alpha_i\)), link offset (\(d_i\)), and joint variable (\(\theta_i\)) were determined for each joint as summarized in Table~\ref{tab:dh_param}. These parameters represent the geometric relationships between adjacent links in the robot’s kinematic chain.  

\begin{table}
\centering
\caption{Denavit-Hartenberg (DH) parameters for Kinova Gen3 7-DOF robotic arm.}
\begin{tabular*}{\columnwidth}{@{\extracolsep{\fill}}ccccc@{}}
\hline
$i$ & $\alpha_i$ (rad) & $a_i$ (m) & $d_i$ (m) & $\theta_i$ (rad) \\
\hline
0 (base) & $\pi$  & 0 & 0 & 0 \\
1 & $\pi/2$ & 0 & -0.2848 & $q_1$ \\
2 & $\pi/2$ & 0 & -0.0118 & $q_2 + \pi$ \\
3 & $\pi/2$ & 0 & -0.4208 & $q_3 + \pi$ \\
4 & $\pi/2$ & 0 & -0.0128 & $q_4 + \pi$ \\
5 & $\pi/2$ & 0 & -0.3143 & $q_5 + \pi$ \\
6 & $\pi/2$ & 0 & 0 & $q_6 + \pi$ \\
7 (end effector) & $\pi$ & 0 & -0.1674 & $q_7 + \pi$ \\
\hline
\end{tabular*}
\label{tab:dh_param}
\end{table}

The DH parameters were used to construct the forward kinematic model describing the relationship between the robot’s joint angles and end-effector position. The forward kinematics solution is obtained using the Homogeneous Transformation Matrix (HTM) formulation~\cite{wen2014study}. Combining the \(3 \times 3\) rotation matrix (\(R^{0}_{n}\)) and the \(3 \times 1\) translation vector (\(p^{0}_{n}\)), the transformation matrix from the base to the \(n^{th}\) joint is defined as:
\begin{equation}
\textbf{H}_n^0 =
\begin{pmatrix}
\textbf{R}_n^0 & \textbf{p}_n^0 \\
\textbf{0} & 1
\end{pmatrix}
\label{eq:HTM1}
\end{equation}
for each joint \(j = 1, \ldots, 7\), the corresponding transformation matrix between adjacent links is expressed as:

\begin{equation}
\label{eq:Hi}
\small
\textbf{H}_j^{j-1} =
\begin{pmatrix}
c(q_j) & -c(\alpha_j)s(q_j) & s(\alpha_j)s(q_j) & a_j c(q_j) \\
s(q_j) & c(\alpha_j)c(q_j) & -s(\alpha_j)c(q_j) & a_j s(q_j) \\
0 & s(\alpha_j) & c(\alpha_j) & d_j \\
0 & 0 & 0 & 1
\end{pmatrix}
\end{equation}
where \(q_j\) denotes the joint actuator angles, and \(s\) and \(c\) represent the sine and cosine functions, respectively. The parameters \(a_j\), \(d_j\), and \(\alpha_j\) are the DH constants for each link, defining the spatial relationship between consecutive joints. The transformation matrix for the end effector (EE) is defined as:

\begin{equation}
\label{eq:H7}
\small
H_7^{EE} =
\begin{pmatrix}
1 & 0 & 0 & 0 \\
0 & -1 & 0 & 0 \\
0 & 0 & -1 & -0.0615 \\
0 & 0 & 0 & 1
\end{pmatrix}
\end{equation}
the total transformation from the robot base to the tool tip is calculated by multiplying the individual transformations for all joints:

\begin{equation}
\label{eq:Htip}
\textbf{H}_{tip}^0 = \textbf{H}_1^0 \textbf{H}_2^1 \textbf{H}_3^2 \textbf{H}_4^3 \textbf{H}_5^4 \textbf{H}_6^5 \textbf{H}_7^6 \textbf{H}_7^{EE}
\end{equation}
the relative pose of Robot 2 with respect to Robot 1 is obtained by computing their transformations with respect to the NDI tracking system. The base transformations are defined as:

\begin{equation}
\label{eq:HNDI1}
H_{b_1}^{\text{NDI}} = H^{\text{NDI}} 
\left( H_{{m_1}}^{{b_1}} \right)^{-1}
\end{equation}

\begin{equation}
\label{eq:HNDI2}
H_{{b_2}}^{\text{NDI}} = H_{{m_2}}^{\text{NDI}} 
\left( H_{{m_2}}^{{b_2}} \right)^{-1}
\end{equation}

\begin{equation}
\label{eq:HR12}
H_{b_2}^{b_1} = 
\left( H_{{b_1}}^{\text{NDI}} \right)^{-1} 
H_{{b_2}}^{\text{NDI}}
\end{equation}
from the resulting matrix \(H_{b_2}^{b_1}\), the relative position of Robot 2 with respect to Robot 1 is determined by extracting the three elements of the final column, representing the translational displacement between the two robot bases.

\subsection{Simulation-based Training Dataset}
\label{simulation_environment}
\begin{figure}
    \centering
    \includegraphics[width=\linewidth]{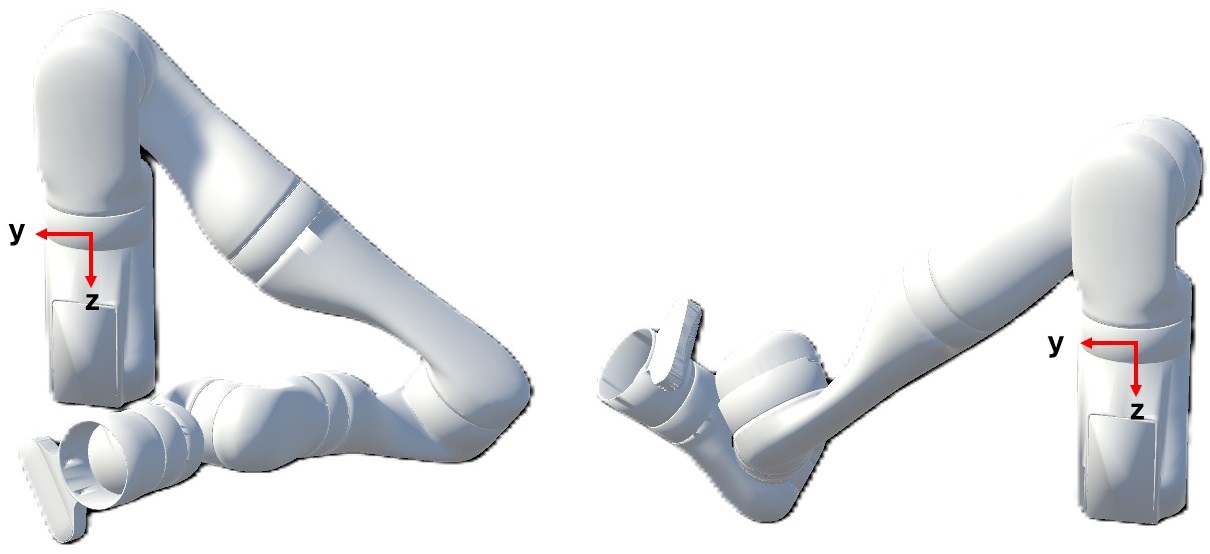}
    \caption{Simulation of two Kinova Gen3 robotic arms in Unity 3D. The environment models the dual-arm setup with accurate link geometries, actuator hierarchies, and kinematic constraints, enabling realistic motion simulation, collision visualization, and data generation for learning-based analysis.}
    \label{fig:Unity_simulation}
\end{figure}

As illustrated in Fig.~\ref{fig:Unity_simulation}, two Kinova Gen3 robotic arms were modeled and simulated within a virtual environment created in Unity 2022.3.32f1. The simulation aimed to replicate the physical structure, kinematic behavior, and motion interactions of the dual-arm setup used in the real-world experiments. The Unified Robot Description Format (URDF) files of the Kinova robots, including mesh representations of each link and joint, were imported into the Unity workspace to construct accurate 3D models of the manipulators. Within the Unity environment, two GameObjects, Robot~1 and Robot~2, were defined, each representing one of the robotic manipulators. Custom C\# scripts were developed to control joint rotations, manage hierarchical link relationships, and define actuator behavior consistent with the robots’ physical constraints. The joint hierarchies were organized according to a parent-child structure, ensuring that each actuator motion propagated correctly through the kinematic chain. The home and zero configurations were defined using data retrieved from the Kinova Application Programming Interface (API), with the zero position corresponding to a configuration where all joint angles were set to zero. The simulation environment enabled the two robots to move through a wide range of configurations and random poses, allowing for systematic testing of motion coordination, workspace overlap, and collision detection within a realistic 3D setting. This setup served as the foundation for generating synthetic data used to train and validate the learning-based collision detection framework described in Section~\ref{Collision Detection Framework}.  

\subsection{Analytical Benchmark}
\label{Collision Detection Algorithms: Analytical}
To detect potential collisions between different configurations of robotic arms, each robotic arm was represented by a series of 1-degree linear Bézier curves. A linear Bézier curve is a parametric representation of a straight line segment defined by its start and end points. This representation simplifies the structural modelling of the robotic arms while maintaining the necessary accuracy for collision detection. Specifically, every link connecting two actuators was represented as a one-dimensional linear Bézier line, where the actuators' positions defined the start and end points of the curve. In this model, each robot link is represented as a straight line segment defined by its start (\(\mathbf{p}_i\)) and end (\(\mathbf{p}_{i+1}\)) points. The equations for the links are parameterized using \(\eta\) and \(\gamma\), where \(\eta, \gamma \in [0, 1]\). Each link for both Robot~1 (\( {}^1\mathbf{l}_{i} \)) and Robot~2 (\( {}^2\mathbf{l}_{i} \)) can be defined using a linear interpolation equation. For Robot~1, the equation for the \(i\)-th link is expressed as: 
\begin{equation}
\label{eq:LR1}
{}^1\mathbf{l}_{i}(\eta) = {}^{1}\mathbf{p}_{i} + \eta \cdot \left({}^{1}\mathbf{p}_{i+1} - {}^{1}\mathbf{p}_{i}\right), \quad \eta \in [0, 1], \quad i = 1, \ldots, 7
\end{equation}
where \({}^{1}\mathbf{p}_{i}\) and \({}^{1}\mathbf{p}_{i+1}\) represent the start and end points of the \(i\)-th link, respectively, and \(\eta\) serves as the interpolation parameter. Similarly, the equation for the \(i\)-th link for Robot~2 is given by:  
\begin{equation}
\label{eq:LR2}
{}^2\mathbf{l}_{i}(\gamma) = {}^{2}\mathbf{p}_{i} + \gamma \cdot \left({}^{2}\mathbf{p}_{i+1} - {}^{2}\mathbf{p}_{i}\right), \quad \gamma \in [0, 1], \quad i = 1, \ldots, 7
\end{equation}
where \(\gamma\) serves as the interpolation parameter, analogous to \(\eta\). The control points \({}^{1}\mathbf{p}_{i}\) and \({}^{2}\mathbf{p}_{i}\) are derived from the homogeneous transformation matrices (\(HTMs\)) corresponding to each joint as indicated in Eq.~\ref{eq:HTM1}. Specifically, for Robot~1, the base of the robot \({}^1\mathbf{p}_{b_1}\) is positioned at the center of the coordinate system, i.e.  
\[
\begin{pmatrix}
0 & 0 & 0
\end{pmatrix}^T
\]
and the rest of the control points are computed using consecutive multiplication of the HTMs:  
\begin{equation}
\label{eq:LHi}
\textbf{H}_{i}^0 = \prod_{j=1}^{i} H_{j}^{j-1}, \quad i=1,\ldots,7
\end{equation}
The position of every point \({}^{1}\mathbf{p}_{i}\) is extracted from the first three elements of the last column of the resulting \(\mathbf{H}_{i}^0\):  
\begin{eqnarray}
\label{eq:PR1}
\mathbf{p}_{b_1} = {}^{1}\mathbf{p}_{0} =
\begin{pmatrix} 
0 \\ 
0 \\ 
0 
\end{pmatrix} \\
{}^{1}\mathbf{p}_{i} = 
\textbf{H}_{i}^0 \textbf{S}
\end{eqnarray}
where:
\begin{equation}
\textbf{S}=
\begin{pmatrix}
0 \\ 0 \\ 0 \\ 1
\end{pmatrix}
\end{equation}
furthermore, for Robot~2, the base of the robot \({}^2\mathbf{p}_{b_2}\) is defined as the first three elements of the last column of \(\mathbf{H}_{b_2}^{b_1}\):  
\begin{eqnarray}
\label{eq:PR2}
\mathbf{p}_{b_2} = {}^{2}\mathbf{p}_{0} = \mathbf{H}_{b_2}^{b_1} \textbf{S} \\
{}^{2}\mathbf{p}_{i} = \textbf{H}_{i}^0 \textbf{S}
\end{eqnarray}
this formulation ensures systematic computation of the positions of all points along the robots' links, based on their respective kinematic chains and the Denavit–Hartenberg (DH) convention. To detect possible collisions, the distance between the two robotic arms was computed by finding the minimum distance between the Bézier curves representing their respective links. Initially, the difference between the link equations was computed as:  
\begin{equation}
\label{eq:dij}
\textbf{d}_{ij}(\eta,\gamma) = {}^1\textbf{l}_{i}(\eta) - {}^2\textbf{l}_{j}(\gamma)
\end{equation}
where \(\textbf{d}_{ij}\) represents the translation vector between an arbitrary point on \({}^1\mathbf{l}_{i}(\eta)\) and a point on \({}^2\mathbf{l}_{j}(\gamma)\). Thus, the size (norm) of \(\textbf{d}_{ij}\) estimates the Euclidean distance (length of the common normal) between the two links of the two robots for a given pair \((\eta,\gamma)\). To obtain the minimum distance between the two robot links, the squared distance function \(\psi\) was analytically derived, differentiated using the chain differentiation rule, and solved for its roots in terms of \((\eta,\gamma)\):  
\begin{eqnarray}
\label{eq:psi}
\psi = \|\textbf{d}_{ij}\|^2 = \textbf{d}_{ij}^\top \textbf{d}_{ij} \\
\label{eq:gradpsi}
\nabla \psi = 
\begin{pmatrix}
\frac{\partial \psi}{\partial \eta}\\
\frac{\partial \psi}{\partial \gamma}
\end{pmatrix} =
\begin{pmatrix}
\frac{\partial \textbf{d}_{ij}^\top}{\partial \eta}\textbf{d}_{ij} + \textbf{d}_{ij}^\top\frac{\partial \textbf{d}_{ij}}{\partial \eta}\\
\frac{\partial \textbf{d}_{ij}^\top}{\partial \gamma}\textbf{d}_{ij} + \textbf{d}_{ij}^\top\frac{\partial \textbf{d}_{ij}}{\partial \gamma}
\end{pmatrix} = \textbf{0}
\end{eqnarray}

These derivatives allow optimization of \(\eta\) and \(\gamma\) to achieve the minimum distance. This process was repeated for all arm configurations of the robots to identify where and between which links the minimum distance occurs, providing a detailed analysis of the collision-free zones and possible interference regions. The optimal \(\eta\) and \(\gamma\) values, obtained by solving the derivative equations, were then substituted back into the link equations to calculate the minimum distance. If this minimum distance was found to be less than 0.2~m, an early collision warning was triggered, consistent with the threshold rationale described in Section~\ref{Performance Metrics}. This hybrid approach offers several advantages. The use of Bézier curves simplifies the collision detection process, avoiding the computational complexity associated with traditional 3D solid geometry models. At the same time, the DH convention ensures precise computation of actuator positions, which is fundamental to accurate collision analysis. Together, these methods provide a robust framework for real-time collision detection in dynamic environments where robotic arms frequently change configurations and operate nearby. Using this approach, the system ensures high efficiency and reliability, making it suitable for complex robotic applications. The linear Bézier representation was adopted as a deliberate modeling choice rather than a coarse approximation. Each Kinova Gen3 link is a slender, approximately straight segment connecting two consecutive rotational joints, so a straight line abstraction captures the dominant geometry of the manipulator while reducing the inter link minimum distance problem to a closed form line segment to line segment problem whose gradient is obtained in closed form (Eqs.~\ref{eq:psi} and~\ref{eq:gradpsi}). This is essential both for generating the large simulation dataset efficiently and for supporting real time inference within the warning layer. The tradeoff is that the physical cross sectional radius of the arms (approximately 75~mm) and the curvature of non-cylindrical link geometries are not explicitly modeled, so the reported distances correspond to axis to axis clearance between idealized line segments rather than to true surface to surface separation. This introduces a systematic offset in configurations where the arms are bulkier or more curved than the line segment approximation assumes and should be considered when interpreting the analytical model in complex geometries.

\subsection{Simulation-Based Collision Detection: Dataset Curation}
\label{Collision Detection Algorithms: Simulation-based}
A custom script was developed within Unity to compute the minimum distance between the simulated robotic arms for each random configuration, enabling real-time collision detection and spatial analysis. To model the individual links of each robotic arm, box colliders were applied to represent the physical geometry of the arm segments, and each collider was assigned to its corresponding link with vertices computed based on their respective positions and orientations in 3D space. The Euclidean distance method was then implemented to determine the shortest distance between the vertices of all possible pairs of box colliders, allowing the identification of the minimum distance between the two robotic arms. The Euclidean distance formula, which calculates the straight-line distance between two points in 3D space, was iteratively applied to all vertex pairs corresponding to the links of the arms, and this process was repeated across all possible configurations to ensure accurate detection of the minimum inter-arm distance for each configuration. This simulation-based approach allowed for continuous and computationally efficient monitoring of spatial proximity, providing a reliable means of validating potential collisions within the virtual environment.

\subsection{Model Training}
\label{Model Training}
\subsubsection{Dataset Preparation}
\label{Dataset Preparation}
The dataset used for training the regressor network was generated through a series of random configurations within the Unity-based simulation. A total of 75,655 configurations were collected by randomly positioning the robotic arms and calculating the minimum distance between them. Each input data point consisted of the following features:
\begin{enumerate}
    \item The relative position vector between the two robots:
    \begin{equation}
        {}^1\mathbf{p}_2 =\mathbf{p}_{b_2} = \begin{pmatrix} p_x \\ p_y \\ p_z \end{pmatrix},
    \end{equation}
    which represents the position of the base of Robot\textendash 2 relative to the base of Robot\textendash 1. Rotational degrees of freedom for the robot bases were omitted as the bases remained fixed throughout the experiments.
    \item Seven joint angles for Robot\textendash 1 and seven joint angles for Robot\textendash 2, corresponding to the specific configurations of their arms at any given moment.
\end{enumerate}

The minimum distance between the two robotic arms was computed in the simulation, and each input data point was formatted as a vector, represented as:
\begin{equation}
\begin{pmatrix}
    p_x & p_y & p_z & q^{(1)}_1 & \dots & q^{(1)}_7 & q^{(2)}_1 & \dots & q^{(2)}_7
\end{pmatrix}^T,
\end{equation}
where \(q^{(1)}_i\) and \(q^{(2)}_i\) denote the joint angles for Robot\textendash 1 and Robot\textendash 2, respectively, and \(d_{min}\) represents the minimum distance between the two robots.
Additionally, the minimum distance values corresponding to the input data were used as the target outputs for this model. Prior to training, several preprocessing steps were applied to prepare the data for input into the NN. First, feature selection was performed to retain the eight most informative joint angles through the permutation based importance analysis described in Section~\ref{Feature Selection}, while the remaining six joint angles and the relative base position were excluded from the network input, as the bases remained fixed throughout the experiments and these features contributed negligible predictive information. Finally, the dataset was randomly divided into two subsets: 90\% of the data was allocated for training and 10\% for validation. This partitioning enabled efficient model learning while allowing for a robust evaluation of generalization performance during the validation phase. The 75,655 configurations were obtained by independently sampling each of the 14 joint angles uniformly within the Kinova Gen3 range of motion, together with the relative base position of Robot~2 with respect to Robot~1, which yields broad coverage of the joint-configuration space for the two-arm system. After the preprocessing and feature-selection steps described above, the resulting dataset was randomly partitioned into 90\% training and 10\% validation, with the held-out split used both to monitor training and to report the regression metrics in Section~\ref{Performance Metrics}. Generalization to unseen configurations is therefore assessed on simulation configurations not seen during training and further probed on the ten physical configurations of Section~\ref{Experimental Validation}, which were not part of the training set.

\subsubsection{Training and Performance}
\label{Training and Performance}
The network was designed as a residual multilayer perceptron (MLP) to regress the minimum distance between the two robotic arms. The input consists of eight joint angles selected through a permutation-based importance analysis, which identified the informative subset of the 14-DOF dual-arm system; the remaining six joints contributed negligible predictive information and were excluded to reduce input noise. Each joint angle \(\theta\) was encoded as \([\sin\theta, \cos\theta]\) to preserve the periodic topology of configuration space, yielding a 16-dimensional input vector. The architecture comprises a linear stem that projects the input to a hidden width of 512 units with batch normalization and GELU activation, followed by eight residual blocks, each containing two linear layers with batch normalization and a skip connection, and a two-layer prediction head that maps through a 64-unit bottleneck to a scalar output passed through a sigmoid function scaled to the observed distance range. These layer widths and depths were selected empirically through trials to balance model capacity and training efficiency. Training used the AdamW optimizer with a cosine annealing learning rate schedule and the Charbonnier loss, which provides smooth gradients near zero residuals. Early stopping with a patience of 60 epochs on the validation loss was employed to prevent overfitting. The dataset was partitioned into training (90\%) and validation (10\%) sets, and the model was evaluated on the held-out validation set using \(R^2\), RMSE, and MAE.

\subsubsection{Feature Selection}
\label{Feature Selection}
To identify the minimum sufficient set of joint inputs, we conducted a permutation importance analysis on the trained baseline model (all 14 joints). For each joint \(j\), the corresponding sin/cos feature pair was randomly permuted in the validation set while all other features remained unchanged, and the resulting drop in \(R^2\) was recorded as the importance score for that joint. Joints J1, J2, J8, and J9, the base and first shoulder joints of each arm, exhibited the largest importance scores (\(\Delta R^2 > 0.67\)), confirming that gross arm positioning dominates the collision distance signal. The mid-arm joints J3, J4, J10, and J11 showed moderate contributions (\(\Delta R^2 \approx 0.03\)--\(0.05\)), while the distal wrist joints J5--J7 and J12--J14 had negligible or slightly negative importance, suggesting they introduce noise rather than signal at the prediction task. A subsequent backward elimination experiment, iteratively retraining after dropping the least important joint, confirmed that retaining the eight joints J1--J4 and J8--J11 preserves within 1\% of the full-14-joint \(R^2\), while using only the four critical joints yields a meaningful drop in accuracy. Accordingly, the final model uses the eight-joint subset as input, balancing sensor parsimony with predictive fidelity.

\section{Validation Studies}
\label{Validation Studies}
The analytical Bézier based minimum distance computation described in Section~\ref{Collision Detection Algorithms: Analytical} and the Unity simulation based pipeline described in Section~\ref{Collision Detection Algorithms: Simulation-based} serve as geometric baselines against which the neural network is evaluated throughout this study. On the ten experimental configurations of Table~\ref{tab:confset}, the analytical baseline exhibits a mean absolute error of 73.3~mm relative to the NDI measurements, and on the held out simulation validation split of 7382 samples, the retrained residual multilayer perceptron exhibits a mean absolute error of 28.7~mm and \(R^2 = 0.940\). The two approaches deliver comparable geometric accuracy, with the neural network offering the additional advantage of a single forward pass at inference time, which is better suited to real time use within the warning layer than the per configuration optimization required by the closed form analytical solution.

\subsection{Simulation-Based Validation}
\label{Simulation-Based Validation}
\subsubsection{Protocol}
\label{Protocol}
The designed analytical algorithm was validated by simulating the robot arms in Unity. Initially, a simulation environment was initiated in Unity, and the URDF files for the Kinova 7-DOF robot were used to model the robots. The control logic was implemented via C\# within the Unity-VS development pipeline. The robots' local rotations were adjusted to control their movement in the Unity environment. To set the relative position and orientation between Robot\textendash 1 and Robot\textendash 2, the transformation matrices for both robots were computed based on their desired relative position and rotation. Using the code in Unity, Robot\textendash 2’s position and rotation were determined by applying a transformation matrix that combined Robot\textendash 1's transformation with the desired relative position and rotation values. This setup ensured that Robot\textendash 2 was placed correctly in relation to Robot\textendash 1.

For generating random configurations, the joint angles and the base position of Robot\textendash 2 were randomized. The random joint angles were generated within the range of motion of Kinova Gen3 robots. For collision detection, BoxColliders were created around each link of the robots. The vertices of the BoxColliders were calculated by transforming the local coordinates of each collider to world coordinates. This method allowed for the precise location of each link’s vertices, which were used to compute the minimum distance between the robots’ colliding links. The Euclidean distance method was employed to find the minimum distance between the vertices of the colliding links. This was done by iterating through all the vertices of Robot\textendash 1 and Robot\textendash 2 and calculating the distance between each pair of vertices. The minimum distance was tracked to identify potential collisions between the robots. The experimental robot configurations were defined later in the paper. These configurations, such as the zero and home positions, were used as reference points to compare real-time robot movements. Future phases of the study will involve physical robots, and the results from the Unity simulation will be verified in real-world conditions, as outlined in later sections of the paper.

\subsubsection{Setup}
\label{Setup}
 The simulation setup was created in Unity 2022.3.32f1, with the project managed through Unity Hub 3.8.0. The robot models used in the simulation are based on the Kinova Gen3 7-DOF robotic arm, chosen for its versatility in robotic manipulation tasks. The models were imported into Unity as URDF files via a custom package obtained from the Unity Package Manager, specifically designed for the Kinova Gen3 robot. Two C\# scripts were written in Visual Studio 2022 to manage the configuration and control of the robots, as well as to compute the minimum distance between the two robots during their interaction. One script controlled the robot's joint angles and generated random configurations, while the other handled the collision detection and distance calculation between the robots. Unity was selected as the simulation platform due to its real-time physics engine, visualization tools, and seamless integration with robotics libraries. The URDF format was used for representing the robot models as it allows easy integration with Unity and accurately defines the robot’s kinematics, including joint positions and link relationships.
 
\subsubsection{Results and Discussion}
\label{Results and Discussion}
The minimum distance values between potentially colliding links were compared between the experimental observations and simulation results. The comparison revealed a fair agreement, with a mean absolute error (MAE) of 73.3 $\pm$ 42.2~mm for the box collider based simulation pipeline relative to the NDI measurements. This demonstrates that the simulation model accurately predicts the minimum distance between the links when the robotic arms are in a collision state, confirming the validity and reliability of the simulation setup. Furthermore, the configurations observed in the simulation closely matched those observed in real-life experiments, further validating the accuracy of the simulated robotic arm behaviours. The observed MAE can be attributed to the simplified representation of the robotic arms in the simulation, where each arm was modelled as a line segment without considering the actual cross sectional radius of the arm (approximately 75~mm). Despite this simplification, the results indicate that the simulation model is reliable for collision detection and for accurately representing robotic arm configurations, making it a valuable tool for predicting real-world performance.

\subsection{Experimental Validation}
\label{Experimental Validation}
\subsubsection{Protocol}
\label{Protocol}
To validate the proposed methodology experimentally, two robotic arms were placed in 10 different random configurations, and one customized passive marker was attached to each base of the robots. Initially, the position values of the base markers ($T_x, T_y, T_z$) and orientation quaternion values ($Q_0, Q_x, Q_y, Q_z$) were captured using the NDI Track application and NDI camera system, a high-precision tool for real-time 3D position tracking. Joint actuator values for each configuration were recorded through the Kinova Web Application. For configurations where a collision occurred between the robotic arms (as determined visually), a calibrated probe was used to mark points on each colliding link. The probe was calibrated using pivot calibration, which involves rotating the tracked tool around a stationary point on the link to determine the 3D position of the probe’s tip and minimize measurement error. Two points were selected on each link in the collision state: the highest reachable point on the link and the lowest reachable point. The position of each point was recorded in real-time as ($T_x, T_y, T_z$) using the NDI camera system. To align with the simulation model, where each robotic arm link was simplified as a line segment, the mean value of the highest and lowest points was calculated to represent the central point of the link. This procedure was repeated for the second robotic arm. The Euclidean distance between the mean points of the colliding links was then computed. This approach ensured consistency between the experimental data and the line-segment representation of the robotic arms used in the simulation model.

\subsubsection{Setup}
\label{Setup}
\begin{figure}
    \centering
    \includegraphics[width=0.8\linewidth]{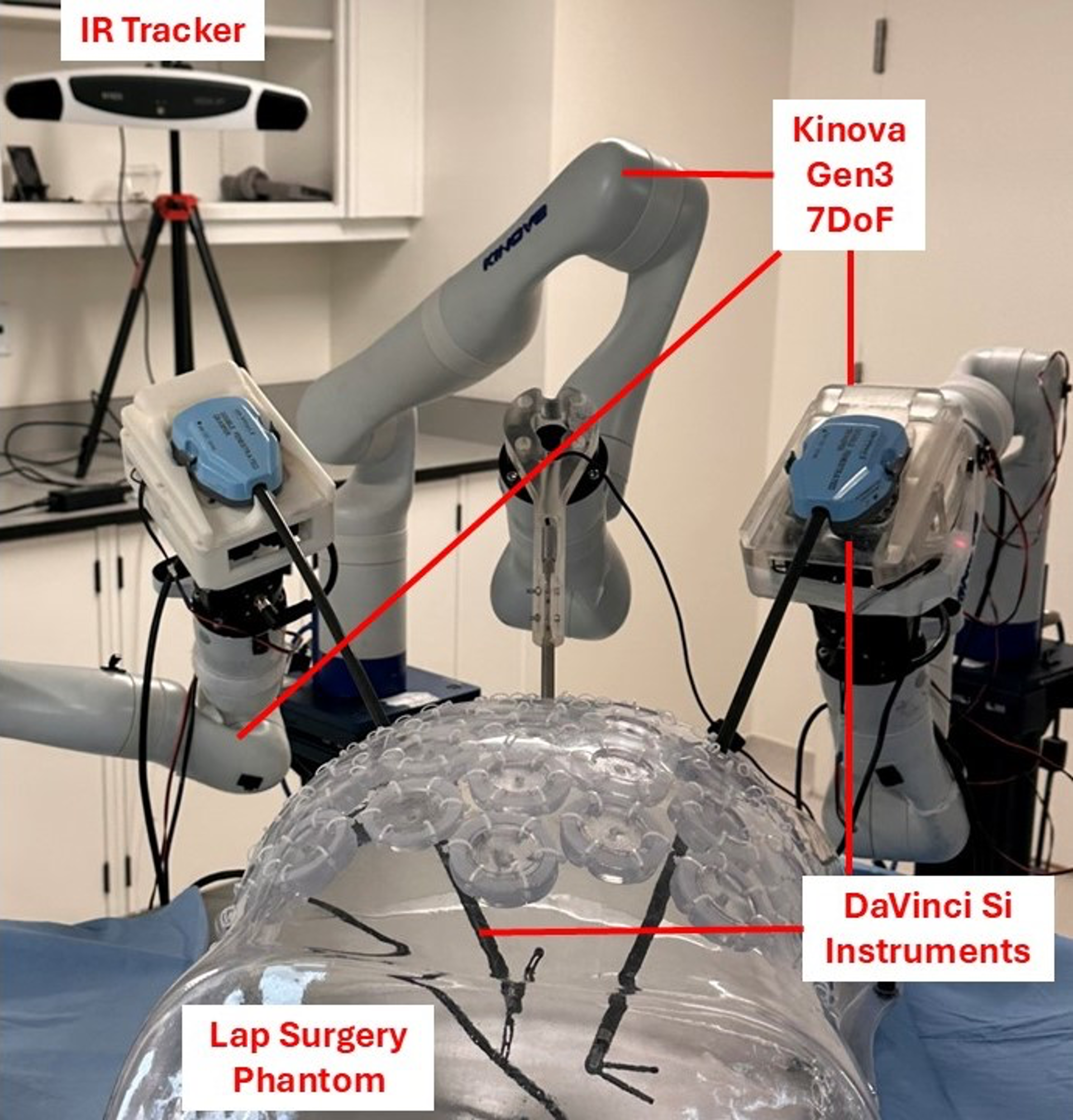}
    \caption{Experimental setup consisting of two Kinova Gen3 7-DoF robotic arms equipped with DaVinci~Si surgical instruments operating around a laparoscopic phantom. The NDI IR tracker was positioned to capture both robots’ base markers for precise motion tracking.}
    \label{fig:setup}
\end{figure}

\begin{figure*}
    \centering
    \begin{subfigure}[t]{0.39\textwidth}
        \centering
        \includegraphics[width=\textwidth]{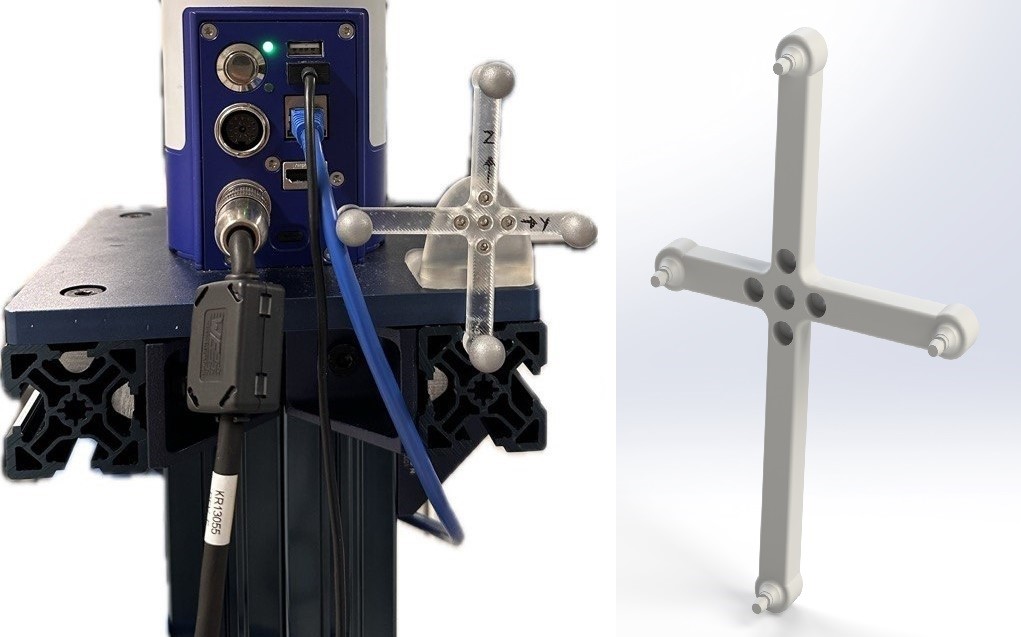}
        \caption{Robot~1 base marker}
    \end{subfigure}
    \hfill
    \begin{subfigure}[t]{0.50\textwidth}
        \centering
        \includegraphics[width=\textwidth]{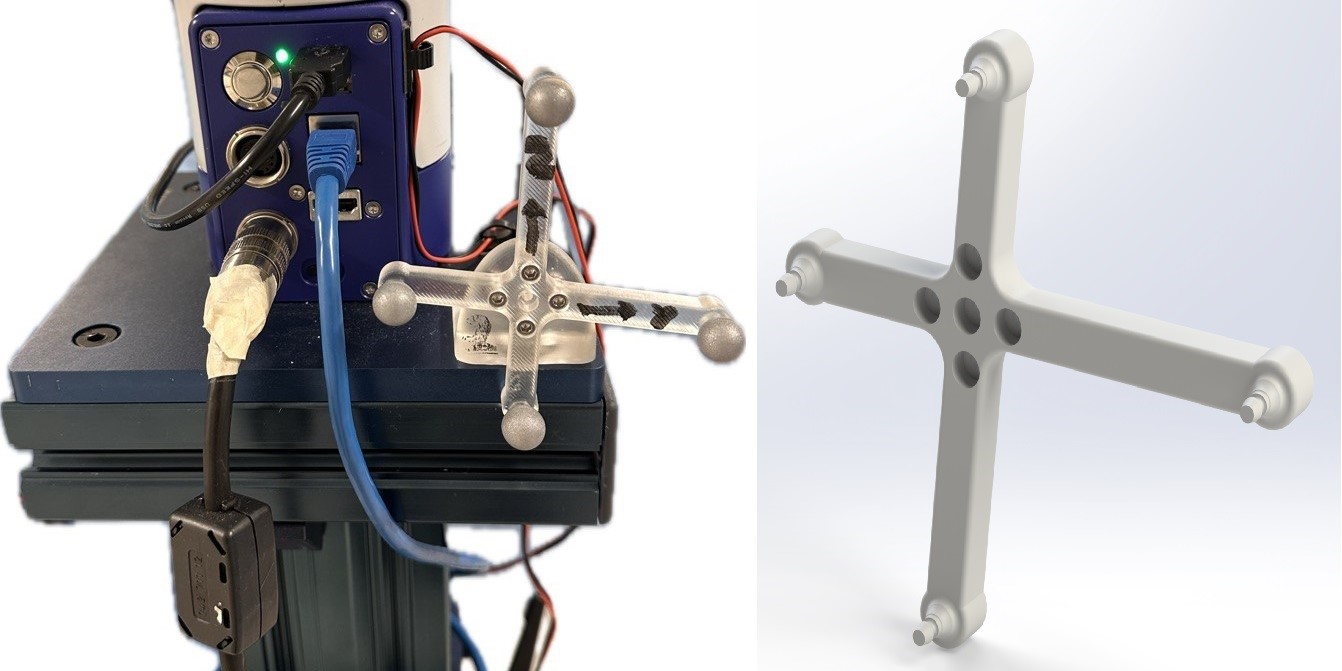}
        \caption{Robot~2 base marker}
    \end{subfigure}
    \caption{Custom-designed passive 4-marker rigid bodies used for NDI optical tracking. 
    (a) Passive 4-marker rigid body attached to the Robot~1 base, with a rendered SOLIDWORKS model illustrating marker design and placement. 
    (b) Passive 4-marker rigid body attached to the Robot~2 base, with its corresponding rendered design. 
    These markers ensured accurate base pose tracking for both robotic arms.}
    \label{fig:base-markers}
\end{figure*}

\begin{figure}
    \centering
    \includegraphics[width=0.8\linewidth]{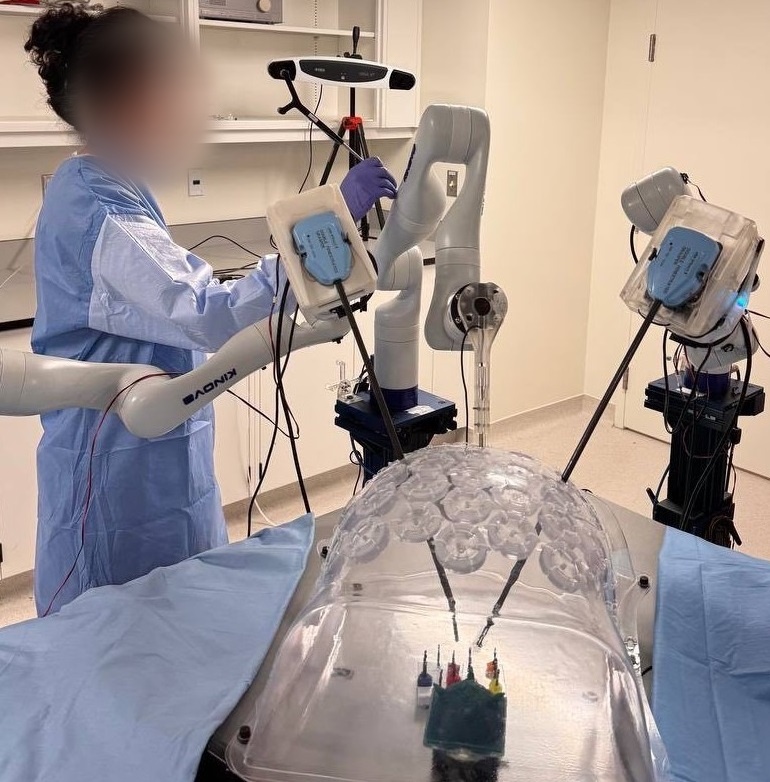}
    \caption{Experimental validation setup showing the dual Kinova Gen3 robotic arms, laparoscopic phantom, and NDI tracking system in operation. The setup enabled real-time tracking, synchronized control, and data acquisition for evaluating the collision detection framework.}
    \label{fig:experiment_setup}
\end{figure}

The experimental setup consisted of two Kinova Gen3 robotic arms, each with 7-DoF, an NDI optical tracking system, a passive 4-marker probe, and two custom-designed passive 4-marker rigid bodies, as shown in Fig.~\ref{fig:setup}. The setup was designed to evaluate the proposed collision detection framework under realistic surgical conditions. Both robotic arms were equipped with DaVinci~Si surgical instruments mounted on their end-effectors, and the system operated around a laparoscopic surgery phantom that mimicked a constrained workspace environment. For Robot~1, a passive 4-marker rigid body was securely attached to its base, with its 3D solid model designed in SOLIDWORKS~2022 following the NDI Tool Design Guide. The design and physical attachment of this marker are illustrated in Fig.~\ref{fig:base-markers}(a). Similarly, for Robot~2, a passive 4-marker rigid body was designed and mounted on its base to provide consistent positional reference during motion, as shown in Fig.~\ref{fig:base-markers}(b). These rigid bodies enabled accurate tracking of the robots’ base poses in real time using the NDI optical tracking system. The NDI infrared (IR) camera was fixed at an elevated position to maintain an unobstructed view of both robots’ base markers throughout the experiment. The setup also included dedicated controllers for each robotic arm, allowing independent or coordinated motion control. The joint actuator values for each random configuration were accessed through the Kinova Web Application, while the positional and rotational data for the probe and rigid body markers were recorded using the NDI Track software (version 6.2, Northern Digital Inc., Waterloo, ON, Canada). The complete experimental validation setup, including the dual-arm configuration, laparoscopic phantom, and tracking system, is shown in Fig.~\ref{fig:experiment_setup}.

\subsubsection{Results and Discussion}
\label{Results and Discussion}
The ten random robot configurations, together with their experimental, analytical, and predicted minimum distances, are summarized in Table~\ref{tab:confset}, and the corresponding joint actuator values are indicated in Table~\ref{tab:RobotConf}. To quantify the rate of error between the minimum distance values obtained from the proposed analytical algorithm and the observed values from the experiments for the 10 random configurations, the absolute error was defined as:
\begin{equation}
Error =  \left| \mathbf{\sqrt{\psi_{exp}}} - \mathbf{\sqrt{\psi_{theo}}} \right|  
\end{equation}
where \( \mathbf{\sqrt{\psi_{exp}}} \) represents the minimum distance values observed and measured using the probe, while \( \mathbf{\sqrt{\psi_{theo}}} \) denotes the theoretical minimum distance values calculated using the analytical method. The results, including the mean error, are reported in Table~\ref{tab:Error}. The mean absolute error of 73.3~mm between the analytical benchmark and the experimental measurements arises from two main sources. The first is parameter uncertainty, including uncertainty in the measured base to base transformation between the two robots, in the pivot calibration of the NDI tracked probe, and in the recorded joint actuator readings. The second is model insufficiency because each robot link is represented as a straight line segment that does not account for the cross sectional radius of the arm or for the curvature of non cylindrical link geometries. Since the analytical model is used to generate training labels, this discrepancy propagates as label noise into the learned model and contributes to the reported regression error. For the present proof of concept the resulting learned accuracy is within an acceptable range for the intended warning layer. Three strategies for reducing this discrepancy can be considered: a higher order Bézier representation that captures the curvature of bulkier link sections, a least squares identification of the kinematic parameters of the analytical model against measured data, and filtering of the NDI tracking input to reduce uncertainty in real time pose measurement.

\begin{table*}
\thisfloatpagestyle{empty}
\centering
\caption{Set of ten robot configurations used for collision simulation with corresponding minimum distances. $\sqrt{\psi_e}$: experimental distance measured using the probe; $\sqrt{\psi_t}$: theoretical distance from the analytical model; $\sqrt{\psi_p}$: predicted distance from the neural network.}
\resizebox{\textwidth}{!}{
\begin{tabular}{c|c|c|c|c}
\hline
\textbf{Case} & \textbf{Configuration} & \boldmath$\sqrt{\psi_{e}}$ (mm) & \boldmath$\sqrt{\psi_{t}}$ (mm) & \boldmath$\sqrt{\psi_{p}}$ (mm) \\ \hline\hline
1 & \includegraphics[width=0.18\textwidth]{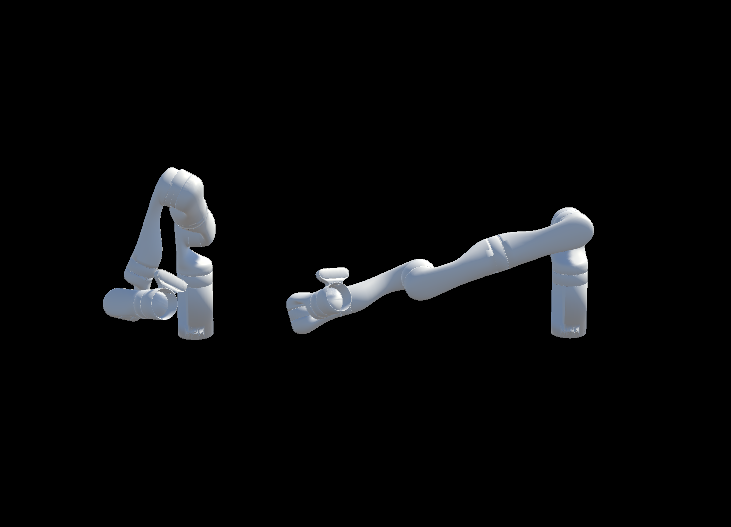} & 305 & 191 & 332 \\ \hline
2 & \includegraphics[width=0.18\textwidth]{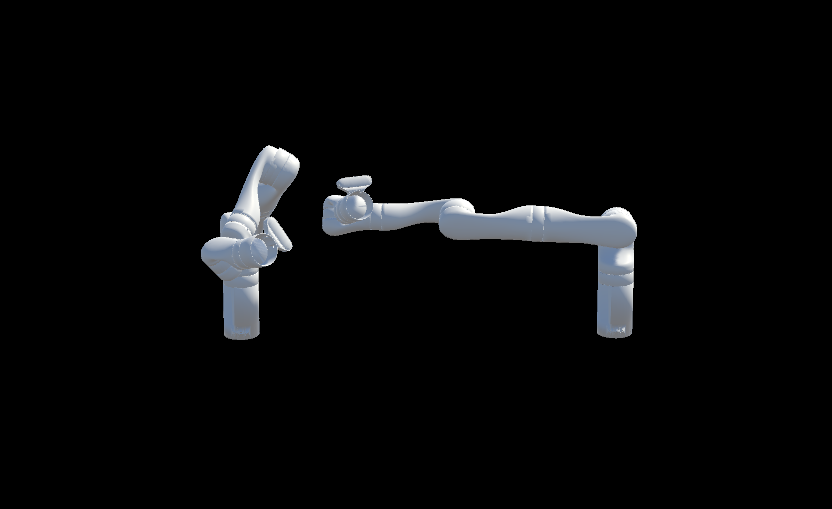} & 178 & 150 & 662 \\ \hline
3 & \includegraphics[width=0.18\textwidth]{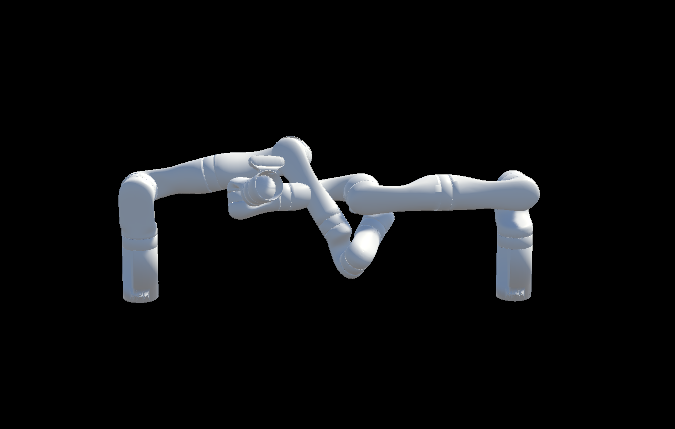} & 64 & 135 & 715 \\ \hline
4 & \includegraphics[width=0.18\textwidth]{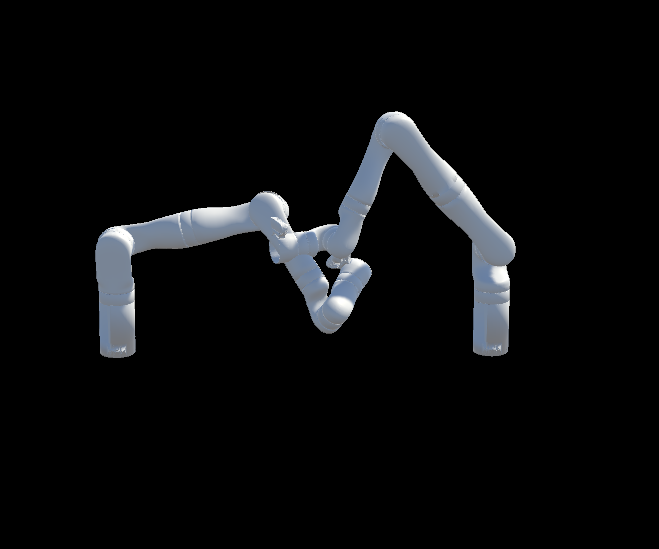} & 65 & 166 & 656 \\ \hline
5 & \includegraphics[width=0.18\textwidth]{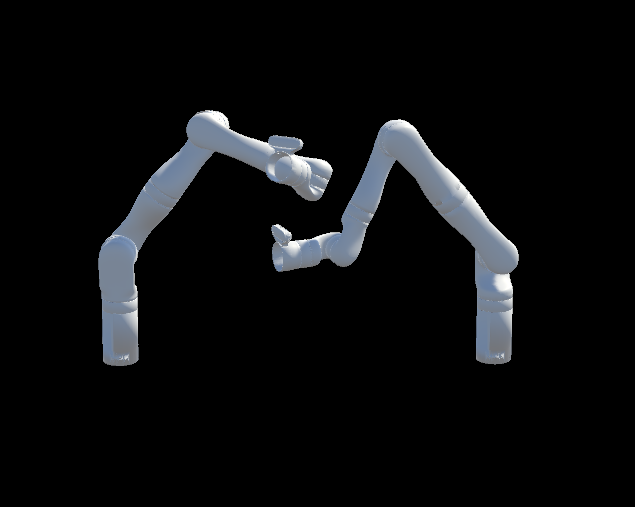} & 115 & 118 & 669 \\ \hline
6 & \includegraphics[width=0.18\textwidth]{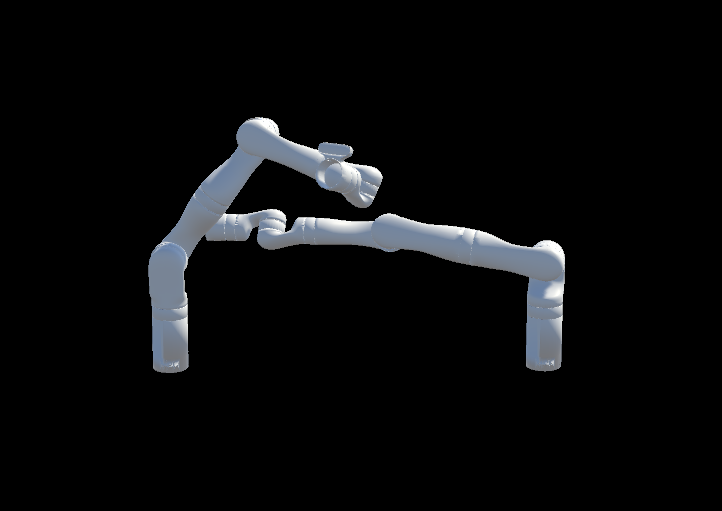} & 163 & 114 & 690 \\ \hline
7 & \includegraphics[width=0.18\textwidth]{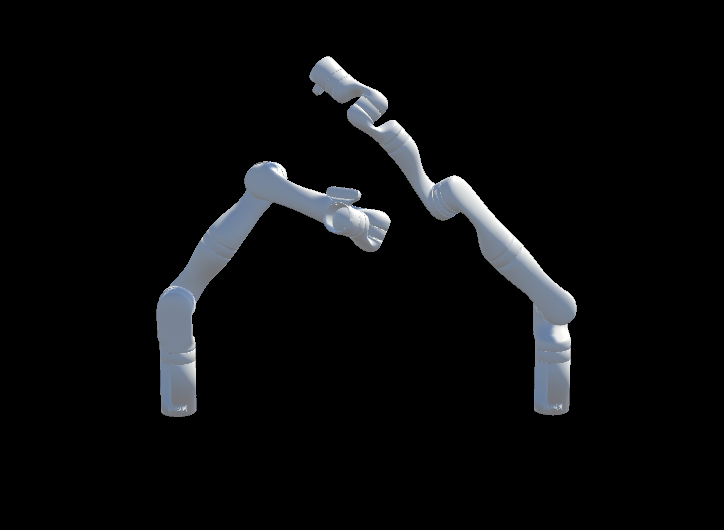} & 141 & 83 & 469 \\ \hline
8 & \includegraphics[width=0.18\textwidth]{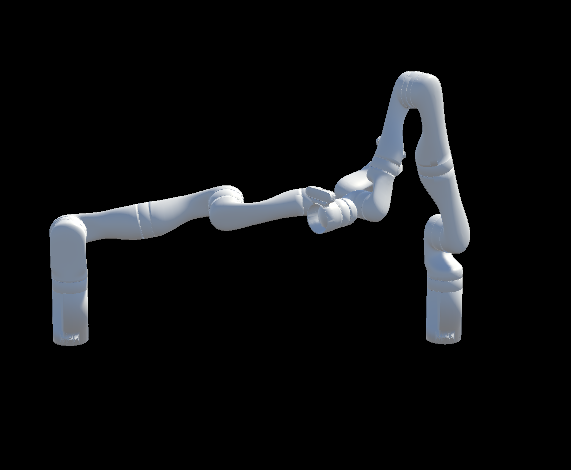} & 191 & 50 & 737 \\ \hline
9 & \includegraphics[width=0.18\textwidth]{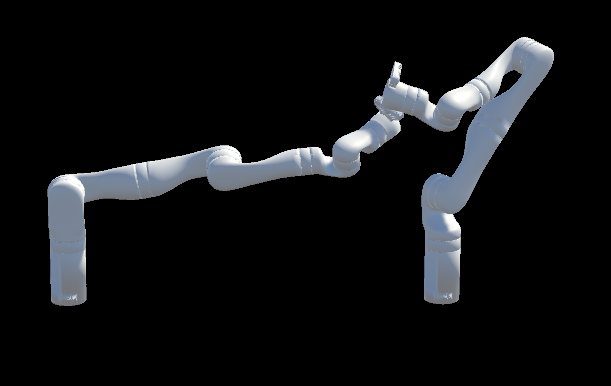} & 110 & 170 & 704 \\ \hline
10 & \includegraphics[width=0.18\textwidth]{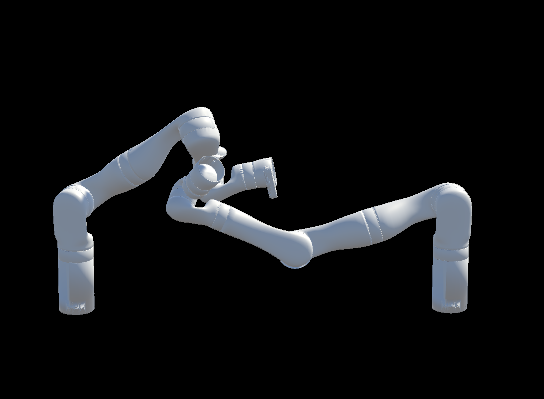} & 140 & 32 & 684 \\ \hline
\end{tabular}}
\label{tab:confset}
\end{table*}

\begin{table*}[t]
\centering
\caption{Joint angles for the set of ten robot configurations in collision.}
\resizebox{\textwidth}{!}{
\begin{tabular}{c|ccccccc|ccccccc}
\toprule
 & \multicolumn{7}{c|}{\textbf{Robot--1}} & \multicolumn{7}{c}{\textbf{Robot--2}} \\ 
\cmidrule(lr){2-8} \cmidrule(lr){9-15}
\textbf{Conf.} & $q_1$ & $q_2$ & $q_3$ & $q_4$ & $q_5$ & $q_6$ & $q_7$ & $q_1$ & $q_2$ & $q_3$ & $q_4$ & $q_5$ & $q_6$ & $q_7$ \\
 & (rad) & (rad) & (rad) & (rad) & (rad) & (rad) & (rad) & (rad) & (rad) & (rad) & (rad) & (rad) & (rad) & (rad) \\
\midrule
1  & 4.88 & 1.84 & 3.36 & 0.06 & 5.40 & 2.10 & 3.02 & 6.25 & 1.04 & 3.62 & 4.28 & 5.48 & 2.06 & 1.34 \\ 
2  & 4.64 & 1.52 & 3.42 & 6.25 & 5.47 & 2.00 & 3.36 & 0.33 & 0.93 & 3.80 & 4.95 & 5.85 & 1.37 & 1.39 \\ 
3  & 4.96 & 1.50 & 3.38 & 6.27 & 5.43 & 1.74 & 3.11 & 1.42 & 1.29 & 0.60 & 5.07 & 5.51 & 2.10 & 1.67 \\ 
4  & 4.92 & 0.69 & 3.13 & 4.32 & 6.11 & 1.08 & 1.65 & 1.42 & 1.29 & 3.60 & 5.07 & 5.51 & 2.10 & 1.67 \\ 
5  & 4.92 & 0.69 & 3.13 & 4.32 & 6.11 & 1.08 & 1.65 & 1.42 & 0.61 & 3.50 & 5.00 & 1.40 & 1.76 & 0.27 \\ 
6  & 4.48 & 1.43 & 2.79 & 6.13 & 5.10 & 0.02 & 3.09 & 1.42 & 0.61 & 3.50 & 5.00 & 1.40 & 1.76 & 0.27 \\ 
7  & 4.91 & 0.75 & 2.30 & 0.05 & 4.02 & 5.61 & 4.73 & 1.42 & 0.61 & 3.50 & 5.00 & 1.40 & 1.76 & 0.27 \\
8  & 5.44 & 0.29 & 3.13 & 3.99 & 0.04 & 1.00 & 1.54 & 1.56 & 1.38 & 2.85 & 6.19 & 4.93 & 0.74 & 3.26 \\ 
9  & 5.23 & 5.60 & 2.83 & 3.74 & 1.15 & 0.98 & 0.37 & 1.56 & 1.38 & 2.85 & 6.19 & 4.93 & 0.74 & 3.26 \\ 
10 & 1.67 & 4.43 & 2.97 & 5.58 & 2.32 & 1.71 & 3.45 & 1.18 & 0.98 & 2.09 & 4.14 & 0.12 & 1.64 & 2.44 \\ 
\bottomrule
\end{tabular}
}
\label{tab:RobotConf}
\end{table*}

\begin{table}
\centering
\caption{Error Analysis of Theoretical and Experimental Minimum Distances.}
\label{tab:Error}
\begin{tabular*}{\columnwidth}{@{\extracolsep{\fill}}c|ccc@{}}
\hline
\# & $\sqrt{\psi_{exp}}$ (mm) & $\sqrt{\psi_{theo}}$ (mm) & $Error$ (mm) \\
\hline
1  & 305 & 191 & 114 \\ 
2  & 178 & 150 & 28 \\ 
3  & 64  & 135 & 71 \\ 
4  & 65  & 166  & 101 \\ 
5  & 115 & 118 & 3 \\ 
6  & 163 & 114 & 49 \\ 
7  & 141 & 83  & 58 \\ 
8  & 191 & 50  & 141 \\ 
9  & 110 & 170 & 60 \\ 
10 & 140 & 32  & 108 \\ 
\hline
Mean & -- & -- & 73.3 \\ 
\hline
\end{tabular*}
\end{table}

\subsection{Validation on Test Data}
\label{NN-based Validation}
\subsubsection{Protocol}
\label{Protocol}
The NN regressor was validated using a systematic protocol to ensure its robustness and generalization capability. The dataset, comprising 75,655 configurations generated through random robotic arm positions in a Unity-based simulation, was randomly partitioned into 90\% training and 10\% held out validation. After removing configurations with invalid or out of range entries during preprocessing, this yielded 7382 held out validation samples that were not seen during training. Each of the 14 joint angles was sampled uniformly within the Kinova Gen3 range of motion, together with the relative base position of Robot~2 with respect to Robot~1, which yields broad coverage of the joint configuration space for the two arm system. Generalization to unseen configurations is therefore assessed on this 10\% held out split and further probed on the ten physical configurations of Section~\ref{Experimental Validation}, none of which were seen during training. Preprocessing steps included feature selection, with eight of the fourteen joint angles retained through a permutation-based importance analysis as described in Section~\ref{Feature Selection}, and sine cosine encoding of each retained joint angle to preserve the periodic topology of configuration space. The architectural, optimization, and loss choices are summarized in Section~\ref{Training and Performance}, and the model was evaluated on the held out validation split using \(R^2\), RMSE, and MAE.

\subsubsection{Setup}
\label{Setup}
The training and validation process was conducted in a controlled computational environment to ensure consistent and reproducible results. The NN was implemented using Python 3.11 and PyTorch 2.2 library, utilizing a workstation with an Apple MPS (Metal Performance Shaders, Apple Silicon) GPU (Apple Inc., Cupertino, CA, USA), 128GB of RAM for efficient computation. The dataset, consisting of 75,655 configurations of robotic arm positions, was preprocessed and loaded using NumPy 1.26 and SciPy 1.11. The training pipeline incorporated extensive debugging and logging tools to monitor the loss and validation performance during training. Random seeds were set for data splitting and model initialization to ensure reproducibility. The AdamW optimizer (PyTorch 2.2 implementation) and cosine annealing scheduler were implemented to dynamically adjust the learning rate based on validation loss, improving training stability. The model was trained over 2500 epochs, with batch processing employed to iterate over the entire dataset in each epoch. The use of the Charbonnier loss function facilitated robust handling of outliers during optimization, further enhancing the reliability of the validation process.

\subsubsection{Results and Discussions}
\label{Results and Discussion}
In this section, we present the results obtained from the trained NN. To assess the network's performance in the context of the experimental measurements, we compared the predicted minimum distances with those recorded during the ten experimental configurations. The comparison is summarized in Table~\ref{tab:comparison}. The mean relative error reported in the final row of Table~\ref{tab:comparison} is computed only over configurations with experimental distance \(\geq 200\)~mm (rows without an asterisk), which lie outside the near contact regime. The five configurations with experimental distances below 200~mm, marked with an asterisk, are excluded from the mean because the trained network is known to overestimate distance in this region, as discussed in Section~\ref{Performance Metrics}; these rows are retained in the table for full transparency. Under this rule the mean relative error over the non near contact configurations is 4.94\%, while the asterisked rows exhibit large relative errors that reflect the class imbalance of the training set and are explicitly identified as a priority direction for future work.

\begin{table} 
	\centering
	\caption{Comparison of predicted minimum distances with experimental values. The mean relative error is computed only over configurations with experimental distance $\geq$ 200~mm (rows without an asterisk); configurations in the near contact regime ($<200$~mm, marked with an asterisk) are excluded from the mean because the trained network is known to overestimate distance in this region, as discussed in Section~\ref{Performance Metrics}.} 
	\begin{tabular*}{\columnwidth}{@{\extracolsep{\fill}}cccc@{}}
		\toprule 
		Configuration & Predicted (mm) & Experimental (mm) & Relative Error (\%) \\
		\midrule 
		1 & 330 & 305 & 8.20 \\
		2 & 662 & 178 & 271.91\textsuperscript{*} \\
		3 & 715 & 642 & 11.37 \\
		4 & 656 & 653 & 0.46 \\
		5 & 669 & 115 & 481.74\textsuperscript{*} \\
		6 & 690 & 163 & 323.31\textsuperscript{*} \\
		7 & 469 & 455 & 3.08 \\
		8 & 737 & 749 & 1.60 \\
		9 & 704 & 110 & 540.00\textsuperscript{*} \\
		10 & 684 & 148 & 362.16\textsuperscript{*} \\ \midrule
		Mean& - & - & 4.94\%\\
		\multicolumn{4}{l}{\textsuperscript{*}: Experimental distance $<200$mm}\\
		\bottomrule
	\end{tabular*}
	\label{tab:comparison} 
\end{table}

\subsection{Performance}
\label{Performance Metrics}
\begin{figure}
    \centering
    \includegraphics[width=\linewidth]{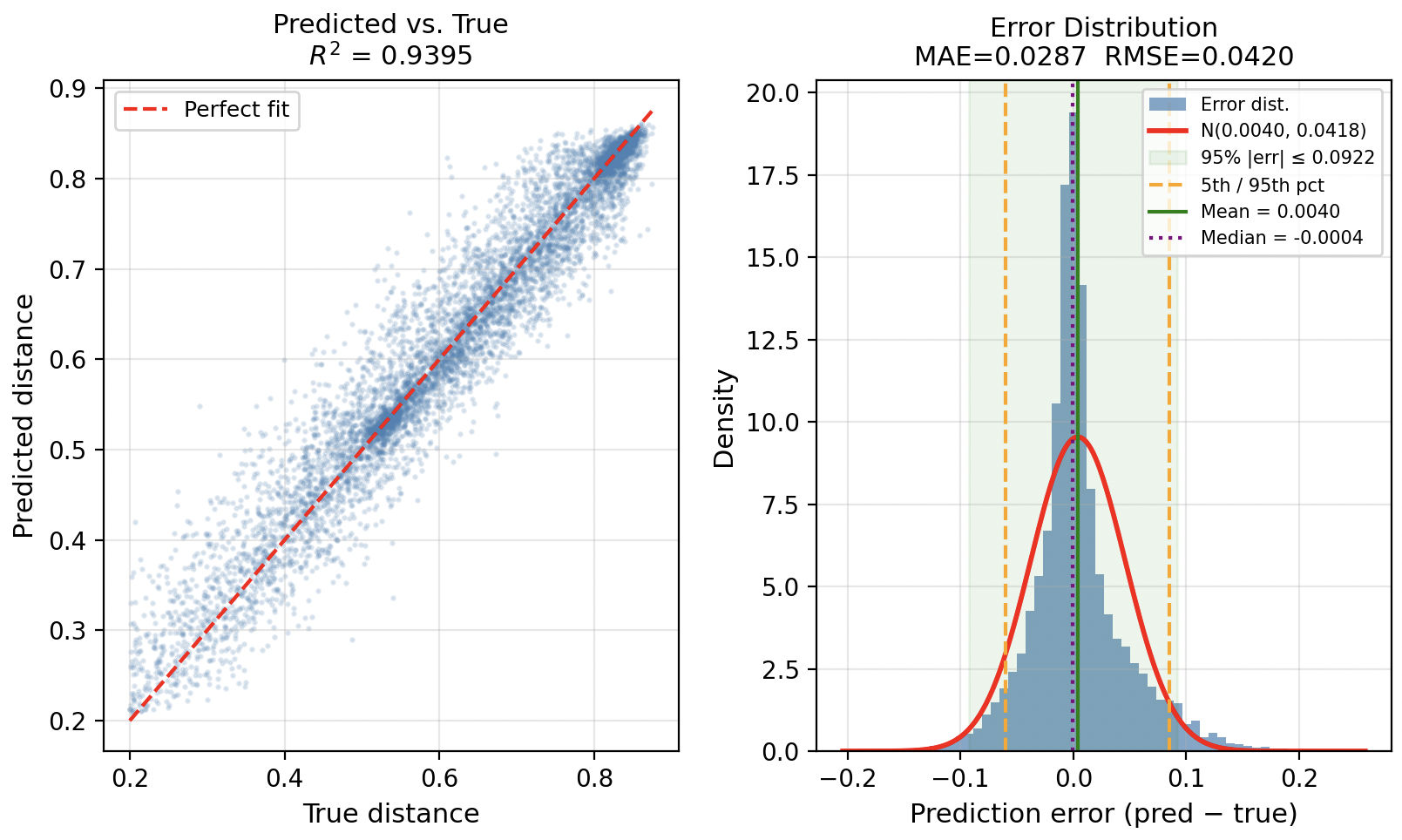}
    \caption{Performance of the retrained residual MLP on the held-out validation set of 7382 samples. Left: predicted minimum distance versus ground truth, with the dashed red line indicating the perfect-fit \(y=x\) direction; the tight clustering of points around this line corresponds to \(R^2 = 0.940\). Right: distribution of prediction errors (predicted minus true), which is approximately Gaussian with mean bias 4.0~mm and median \(-0.4\)~mm. Vertical dashed lines mark the 5th and 95th percentiles, with 95\% of errors falling within \(\pm 92.2\)~mm, confirming that the revised model generalizes consistently across the workspace without systematic bias. The MAE and RMSE values are 28.7~mm and 42.0~mm, respectively.}
    \label{fig:T1}
\end{figure}

The network’s performance was evaluated using several key metrics, including the MAE, root mean squared error (RMSE), and coefficient of determination (\(R^2\)). The performance metrics are summarized in Table~\ref{tab:metrics}, where the model’s predictions were compared with the actual values obtained from the experiments. The mean absolute error must be interpreted in the context of the Kinova Gen3 geometry and the intended role of the framework as an early collision warning layer. The arm can be approximated as a cylinder of radius 75~mm, so a center-line-to-center-line distance of 150~mm corresponds to surface-to-surface contact between two adjacent arms. The 200~mm warning threshold was therefore deliberately selected so that a warning is triggered when the surface-to-surface clearance falls to approximately 50~mm, providing a safety margin that allows the teleoperation system to arrest motion before the arms are driven into collision. The reported error quantifies center-line-to-center-line deviation across the full workspace rather than contact precision, and the framework is intended to complement, rather than replace, precise low-level control. On a held-out validation split of 7382 samples, the revised network achieved \(R^2 = 0.940\), RMSE \(= 42.0\)~mm, MAE \(= 28.7\)~mm, and a near-zero mean bias of 4.0~mm, with 95\% of errors within \(\pm 92.2\)~mm and 99\% within \(\pm 134\)~mm, as summarized in Table~\ref{tab:metrics} and illustrated in Fig.~\ref{fig:T1}. The error distribution is approximately Gaussian and centered near zero, confirming the absence of systematic bias.

To further assess the framework from a classification perspective, the 0.2 m warning threshold was applied. The 200 mm threshold was anchored in the geometry of the Kinova Gen3 arm, which can be approximated as a cylinder of cross-sectional radius 75 mm, so a centre-line-to-centre-line distance of 150 mm corresponds to surface-to-surface contact and the 200 mm threshold provides an additional 50 mm clearance margin that allows the teleoperation system to arrest motion before contact.

Treating predictions below 200 mm as collision warnings and held-out validation distances below 200 mm as ground-truth collisions, the balanced model evaluated on the 7382-sample validation set (304 true collision instances) yields TP~\(= 302\), FP~\(= 19\), FN~\(= 2\), TN~\(= 7359\), giving a recall of 99.3\%, a precision of 94.1\%, a specificity of 99.7\%, a false positive rate of 0.3\%, and an F1-score of 0.966. This represents a qualitative shift from the original TP~\(= 0\) outcome and confirms that targeted near-collision sampling, combined with weighted training and an extended output range, resolves the class-imbalance failure mode identified above.

\begin{table}
\centering
\caption{Performance metrics for the trained NN on the held-out validation set (7382 samples).}
\begin{tabular*}{\columnwidth}{@{\extracolsep{\fill}}c|c@{}}
\toprule
Metric & Value \\
\midrule
Mean Absolute Error (MAE) & 28.7~mm \\
Root Mean Squared Error (RMSE) & 42.0~mm \\
Mean bias & 4.0~mm \\
Median error & $-0.4$~mm \\
95\% error bound ($|e|$) & 92.2~mm \\
R-squared (\(R^2\)) & 0.940 \\
\bottomrule
\end{tabular*}
\label{tab:metrics}
\end{table}

\begin{figure}
\centering 
\includegraphics[width=\linewidth]{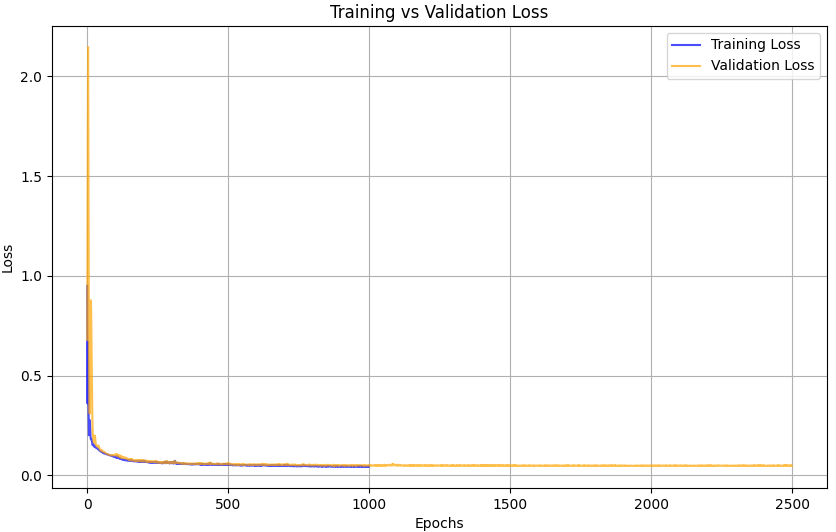} 
\caption{Training and validation loss curves of the NN over 2500 epochs. The rapid convergence within the first few hundred epochs, followed by stable loss values, indicates effective learning and minimal overfitting, confirming the network’s robustness and generalization capability.} 
\label{fig:T2} 
\end{figure}

To evaluate the learning behaviour of the NN during training, the loss values for both the training and validation datasets were recorded and plotted for each epoch, as shown in Fig.~\ref{fig:T2}. This plot provides insights into the convergence of the model and its ability to generalize to unseen data. It demonstrates that the training loss steadily decreases as the epochs progress, indicating that the model successfully optimizes the objective function. The validation loss follows a similar trend, stabilizing after approximately 15 epochs. This stabilization suggests that the model achieves a balance between fitting the training data and generalizing to the validation data without significant overfitting.

The retrained residual MLP demonstrated predictive performance sufficient for early collision warning based on the eight selected joint angles, with a validation \(R^2\) of 0.940 and consistent agreement along the identity direction. The reported MAE quantifies the center line to center line distance between the two manipulators modeled as line segments, rather than the surface to surface clearance between their physical bodies, since the Bézier representation does not account for the 75~mm cross sectional radius of the Kinova Gen3 arm. The framework is therefore intended as an early collision warning layer, flagging configurations below the 0.2~m threshold, rather than as a fine positioning controller. A clear limitation remains in the near contact regime: For the configurations in Table~\ref{tab:comparison} with experimental distances below 200~mm, the network tended to overestimate the minimum distance, resulting in large relative errors in the safety critical region. This behavior reflects the class imbalance of the training set, in which near collision configurations are under represented among the 75,655 randomly sampled poses. Addressing this limitation through targeted sampling of near contact configurations, a weighted loss that emphasizes small distances, and a surface aware distance formulation that incorporates the arm radius is identified as a priority direction for future work to improve reliability in the safety critical regime.

\section{Conclusions}
\label{Conclusion and Future Work}
This study presented an integrated approach to minimum distance estimation and collision-aware warning for multi-arm robotic systems through analytical modeling, simulation, and learning from simulation methods. The proposed Bézier-based analytical framework provided an efficient geometric representation for modeling robotic links and enabled the computation of inter-arm distances with high precision. The Unity-based simulation platform further allowed the exploration of diverse arm configurations and validation of proximity estimation models under realistic kinematic conditions. Moreover, the deep learning model trained on the synthetic dataset demonstrated strong predictive accuracy for estimating minimum distances in real time, supporting safe and coordinated operation of robotic manipulators in shared workspaces. While the framework achieved robust performance, several directions remain for future work. First, incorporating more complex collision geometries and surface contact models would improve the accuracy of proximity estimation, particularly for robots with non-cylindrical or flexible links. Second, extending the analytical formulation to include dynamic modeling and time-varying constraints would enable predictive collision avoidance rather than post-detection estimation. Third, integrating additional sensory feedback such as force or tactile data could enhance situational awareness and support active control strategies for collision avoidance. Finally, deploying the trained neural model on real robotic hardware for closed-loop validation will further verify the framework’s generalizability and real-time performance in clinical and industrial environments. Together, these extensions will advance the presented framework toward a complete, adaptive, and learning-based safety system for next-generation surgical and collaborative robotic platforms.

\section*{{Funding}}
This work was supported by Montreal General Hospital Foundation and Natural Science and Engineering Research Council (NSERC) of Canada.

\section*{{Conflict of Interest/Competing Interests}}
The authors declare no conflicts of interest.

\section*{{Ethics Approval and Consent to Participate}}
Not applicable. This study did not involve human participants or animals; all experiments used robotic manipulators and an inanimate laparoscopic phantom.

\section*{{Data Availability}}
The original contributions presented in this study are included in the article. Further inquiries can be directed to the corresponding authors.

\section*{{Author Contributions}}
S. Ghiasi: Investigation; Data curation; Formal analysis; Methodology; Writing (original draft).
M. Roshanfar: Investigation; Data curation; Formal analysis; Visualization; Writing (original draft).
J. Barralet: Conceptualization; Methodology; Writing (review \& editing); Supervision.
L. Feldman: Conceptualization; Writing (review \& editing); Supervision.
A. Hooshiar: Conceptualization; Writing (review \& editing); Funding acquisition; Supervision. All authors have read and agreed to the published version of the manuscript.

\bibliography{sn-bibliography}
\end{document}